\pdfoutput=1
\documentclass[11pt]{article}

\usepackage[final]{acl}

\usepackage{times}
\usepackage{latexsym}
\usepackage{amssymb}
\usepackage{array}
\usepackage{arydshln}
\usepackage{booktabs}
\usepackage{graphicx}
\usepackage{amsmath}
\usepackage{algorithm}
\usepackage{algpseudocode}
\usepackage{multirow}
\usepackage{multicol}
\usepackage{longtable}
\usepackage{setspace}

\usepackage[russian,english]{babel}
\DeclareUnicodeCharacter{0301}{\'{e}}

\usepackage[T1]{fontenc}

\usepackage[utf8]{inputenc}
\usepackage{microtype}
\usepackage{inconsolata}
\usepackage{graphicx}

\title{\textit{Writing Like the Best}: Exemplar-Based Expository Text Generation}

\author{
    Yuxiang Liu \quad Kevin Chen-Chuan Chang \\
    University of Illinois at Urbana-Champaign, USA \\
    \texttt{\{yuxiang, kcchang\}@illinois.edu}
}

\begin{document}
\selectlanguage{english}

\maketitle
\begin{abstract}
We introduce the \textit{Exemplar-Based Expository Text Generation} task, aiming to generate an expository text on a new topic using an exemplar on a similar topic. 
Current methods fall short due to their reliance on extensive exemplar data, difficulty in adapting topic-specific content, and issues with long-text coherence. 
To address these challenges, we propose the concept of \textit{Adaptive Imitation} and present a novel \textsc{Recurrent Plan-then-Adapt} (\textsc{\textbf{RePA}}) framework. 
\textsc{RePA} leverages large language models (LLMs) for effective adaptive imitation through a fine-grained plan-then-adapt process.
\textsc{RePA} also enables recurrent segment-by-segment imitation, supported by two memory structures that enhance input clarity and output coherence. 
We also develop task-specific evaluation metrics--\textit{imitativeness}, \textit{adaptiveness}, and \textit{adaptive-imitativeness}--using LLMs as evaluators. 
Experimental results across our collected three diverse datasets demonstrate that \textsc{RePA} surpasses existing baselines in producing factual, consistent, and relevant texts for this task. 
\end{abstract}

\section{Introduction}

The increasing demand for digital content creation necessitates advanced natural language generation techniques to produce high-quality text at scale~\cite{gatt2018survey, xu2024towards}. 
%
A system that can reliably generate \textit{expository texts}\footnote{Expository text is writing that presents factual information with the purpose of informing, explaining, or describing a topic for readers, rather than entertaining or persuading them.}
that introduce or summarize topics--such as concepts, entities, or subjects--is highly desirable~\cite{jiang-etal-2024-unknown, shao-etal-2024-assisting}.
This is particularly valuable in domains requiring \textit{consistent} style, like faculty profiles, university overviews, product descriptions, or event introductions, where large volumes of expository texts must be produced from limited examples.
Such a system can significantly reduce human effort by automating the writing process, enhance consistency by maintaining a uniform style, and enable scalability by rapidly generating large volumes of text. 

\begin{figure}[tp!]
\centerline{\includegraphics[width=1\linewidth]{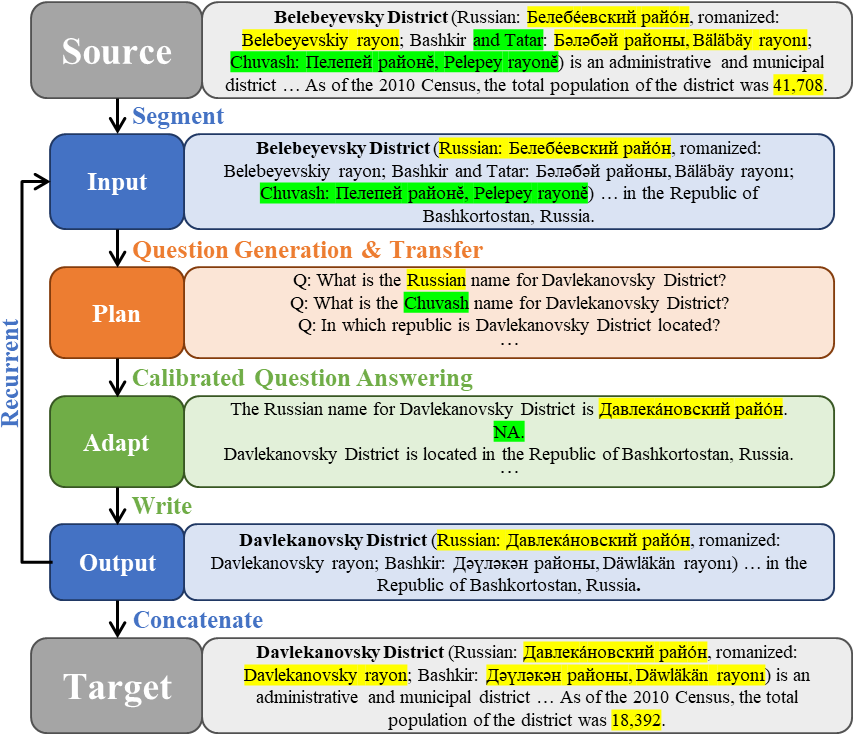}}
\vspace{-0.05in}
\caption{An illustration of \textsc{RePA} for \emph{Exemplar-Based Expository Text Generation}, where yellow text indicates adapted facts and green text indicates discarded facts.}
\label{fig:illustration}
\vspace{-0.15in}
\end{figure}

For expository text generation, prior methods that rely on extensive domain corpora~\cite{balepur-etal-2023-expository, shao2024assisting} struggle when such data is scarce, while approaches for open-ended generation without domain-specific data~\cite{yang-etal-2022-re3, yang-etal-2023-doc} often fail to produce structurally consistent content that adheres to domain conventions.
To this end, we introduce \emph{Exemplar-Based Expository Text Generation}, a novel task that generates an expository text on a new topic by leveraging an exemplar from a similar topic, avoiding reliance on large corpora or unconstrained generation methods.
As shown in Figure~\ref{fig:illustration}, given a text on "Belebeyevsky District", the task is to generate a text on "Davlekanovsky District", preserving the source's overall structure while incorporating topic-specific details.
Since expository texts are typically long-form\footnote{Long-form text refers to content that extends beyond the sentence level, such as one or more paragraphs or even longer passages~\cite{shen-etal-2019-towards, guan-etal-2021-long, hu-etal-2022-planet, min-etal-2023-factscore, liang-etal-2023-open}.}, this task focuses on generating informative, topic-centric long-form content to better address practical needs.

The goal is twofold: to preserve the structure and content of the exemplar, ensuring \textit{cross-topic consistency}, while maintaining accuracy and relevance to the new topic by addressing \textit{cross-topic variability}.
However, existing methods that rely on extensive exemplar data are impractical~\cite{balepur-etal-2023-expository, shao2024assisting}, making it difficult to achieve \textit{cross-topic consistency} due to limited data for generalization.
Additionally, maintaining structural consistency with the exemplar while ensuring factual accuracy for the new topic is particularly challenging due to \textit{cross-topic variability}, which often leads to hallucinations and inaccuracies~\cite{ji2023survey, rawte2023survey}.

To address these challenges, we introduce a new concept called \textbf{Adaptive Imitation}, inspired by how humans learn to write through studying exemplars~\cite{vijayakumar2024exemplification, chen2024exploring, carter2018students, wu2019understanding, wette2014teachers}.
This involves \textit{imitating} the organizational structure of an exemplar while \textit{adapting} the content to fit specific topics and writing requirements.
Specifically, we propose a \textit{plan-then-adapt} approach for fine-grained step-by-step control of LLMs with an \textbf{imitative planner} and an \textbf{adaptive generator}.
The imitative planner leverages LLMs and generates topic-centric outlines framed as questions based on the exemplar. These outlines are then transferred to the new topic through straightforward topic token substitution, ensuring \textit{cross-topic consistency} (\textsc{Plan}: Section~\ref{sec:plan}).
The adaptive generator incorporates retrieval augmentation~\cite{ram-etal-2023-context, shi2023replug} and confidence calibration~\cite{xiong2023can, tian-etal-2023-just} to realize outlines by answering questions factually and discriminatively, effectively adapting topic-specific content and thus addressing \textit{cross-topic variability} (\textsc{Adapt}: Section~\ref{sec:adapt}).

Moreover, to scale to longer texts, we adopt a recurrent, segment-by-segment processing strategy, where both input and output are handled incrementally.
To address coreference issues between input segments, we introduce a \textit{short-term memory} that retains recent input segments during the planning phase.
To reduce output redundancy and maintain coherence, we incorporate a \textit{long-term memory} that summarizes all previously generated segments during the adaptation phase.
This recurrent framework, integrated with the two memory structures, enables our model to process arbitrarily long texts while preserving information integrity.
In summary, we propose a novel \textsc{Recurrent Plan-then-Adapt} (\textsc{\textbf{RePA}}) framework, which integrates planning, adaptation, and recurrent processing with the usage of both short-term and long-term memory.

Finally, as established metrics do not adequately assess stylistic fidelity and knowledge transfer in imitation and adaptation, failing to capture the nuances of this task, we additionally develop task-specific metrics \textit{imitativeness}, \textit{adaptiveness}, and \textit{adaptive-imitativeness} using LLM-as-a-Judge~\cite{zheng2024judging}. Human evaluations further confirm that the LLM-based judgments align well with human judgments. 
We collect three diverse datasets covering both open-domain and domain-specific scenarios. To evaluate our approach, we conduct extensive experiments, including comparisons with strong baselines and ablation studies. The results show that \textsc{RePA} significantly outperforms the baselines by producing factual texts that achieve cross-topic consistency and handle cross-topic variability, with high level of imitativeness and adaptiveness. Additionally, our analysis reveals that each module within RePA contributes effectively to its overall performance.

Our contributions can be summarized as follows:

\noindent (1) We are the first to study \textit{Exemplar-Based Expository Text Generation} task, addressing a practical yet under-explored area with broad applications.

\noindent (2) We present a novel \textsc{Recurrent Plan-then-Adapt} (\textsc{\textbf{RePA}}) framework with two memory structures, achieving fine-grained control of LLMs. 

\noindent (3) We develop task-specific metrics \textit{imitativeness}, \textit{adaptiveness}, and \textit{adaptive-imitativeness} using LLM-as-a-Judge for comprehensive evaluation. 

\noindent (4) We collect three diverse datasets and our extensive experiments demonstrate the superior performance of our proposed method for this task\footnote{Our codes and datasets are available in our repository: \url{https://github.com/liuyuxiang512/RePA.git}.}. 

\section{Related Work}\label{sec:related}

\subsection{Long Input Processing}

Our task is distinguished by its long input. Previous work on long input processing has focused on tasks like outline generation using hierarchical decoders~\cite{zhang2019outline}, summarization with extract-then-generate~\cite{mao-etal-2022-dyle}, divide-and-conquer~\cite{zhang-etal-2022-summn}, graph-based methods~\cite{hua-etal-2023-improving}, and dialogue response generation using retrieval~\cite{kumariretrieval} or memory augmentation~\cite{lu2023memochat, lee-etal-2023-prompted, wang2023recursively}. Some works aim to improve efficiency with sparse attention~\cite{beltagy2020longformer, ivgi-etal-2023-efficient, jin2024llm}. However, these methods primarily generate outputs that are significantly shorter and structurally different from their inputs, containing only a subset of the input's information. In contrast, our task requires outputs comparable in length and structure to the inputs, while incorporating additional relevant knowledge.
%

\subsection{Long-Form Text Generation}

Prior research on long-form text generation (LFTG) ~\cite{koksal2023longform, you2023eipe, zhou2023recurrentgpt, liang-etal-2023-open, adewoyin-etal-2022-rstgen} has addressed tasks such as story~\cite{yang-etal-2022-re3, yang-etal-2023-doc}, data-to-text~\cite{moryossef-etal-2019-step, bai2021infobox}, script~\cite{mirowski2023co}, and expository text generation~\cite{balepur-etal-2023-expository}, where \emph{plan-then-generate} framework is commonly applied to improve coherence of generating long-form text in one go. Various formats of "plan" have been proposed, including key phrases~\cite{hu-etal-2022-planet}, events~\cite{goldfarb-tarrant-etal-2020-content}, data records~\cite{moryossef-etal-2019-step}, and section titles~\cite{shao2024assisting}. Some approaches incorporate discourse guidance~\cite{adewoyin-etal-2022-rstgen} or combine planning with discourse for improved fluency~\cite{sun-etal-2022-summarize}. Other methods enhance planning with retrieval to address logic conflicts~\cite{guan-etal-2020-knowledge} or leverage memory to retain key information~\cite{zhou2023recurrentgpt}.
Unlike existing LFTG methods that use brief inputs and favor open-ended, creative generation with minimal constraints, our task focuses on producing factual, informative outputs from long, detailed inputs, which requires strict adherence to the source text and limits open-endedness.
Additionally, unlike methods that rely on an external or learned "bank of plans", our task necessitates deriving the plan exclusively from a single, lengthy input, ensuring alignment with its content and structure.
%



\subsection{Confidence Calibration of LLMs  }

Confidence calibration--the ability to produce accurate confidence scores indicating the correctness of generated text--is crucial for the trustworthiness of real-world systems. Previous methods required white-box access to model architectures or fine-tuning~\cite{kadavath2022language, jiang-etal-2021-know, lin2022teaching, yang2023alignment, zhang2023r, slobodkin-etal-2023-curious}, which is infeasible for proprietary Large Language Model (LLM) APIs. Recently, \citet{xiong2023can} and \citet{tian-etal-2023-just} explored black-box LLM uncertainty estimation in reasoning and factual short-answer, finding that verbalized confidences emitted as output tokens are typically better calibrated than the model’s conditional probabilities. Building on this, we integrate confidence calibration into our retrieval-augmented generation (RAG) process, allowing for explicit assessment of the accuracy of generated texts.
\section{Method}\label{sec:method}

\begin{figure*}[tp!]
\centerline{\includegraphics[width=1\linewidth]{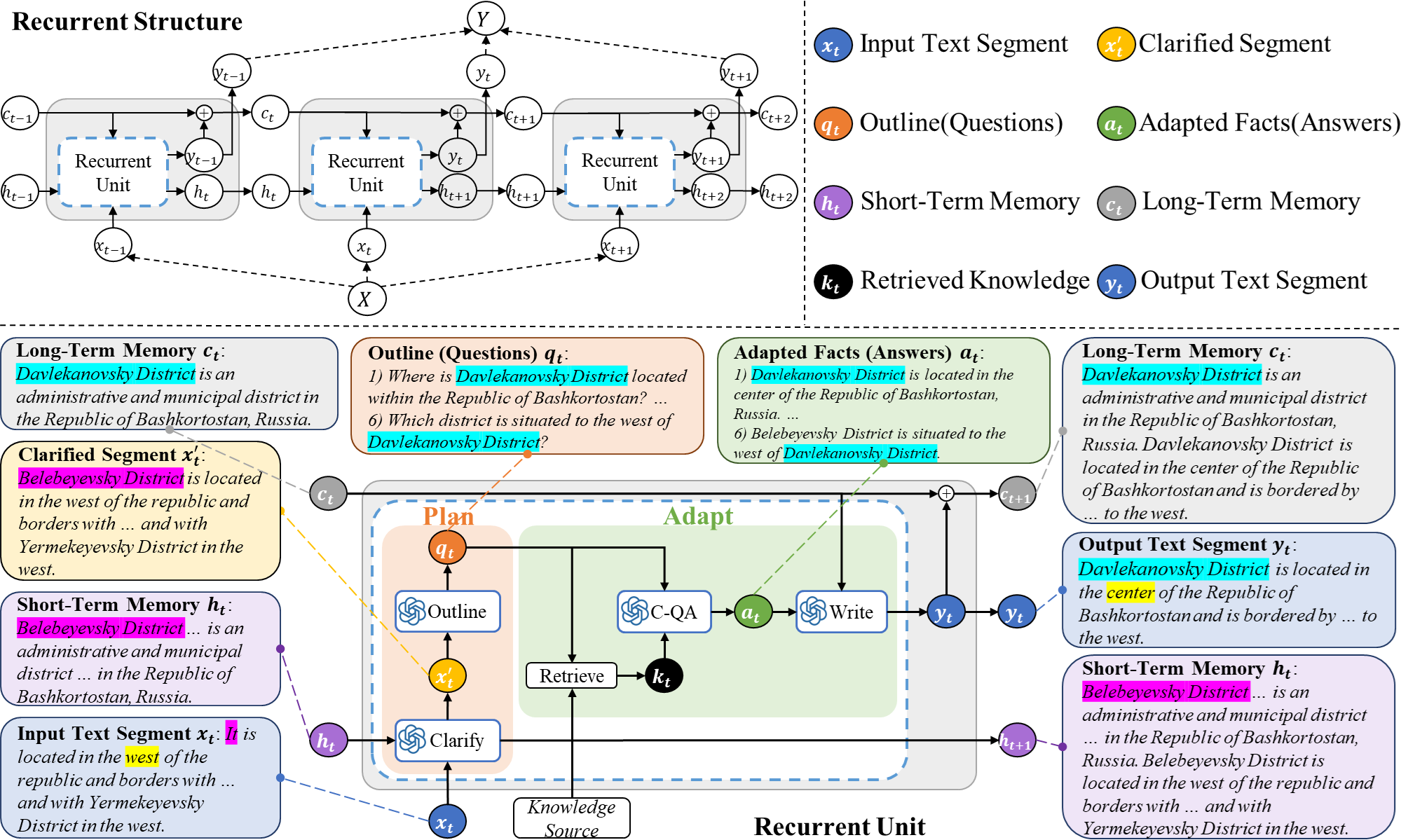}}
\caption{Overview of \textsc{RePA}. Top left shows the high-level recurrent structure for sequential processing. Bottom details the recurrent unit with a running example: "Clarify" and "Outline" in \textsc{Plan}, and "Calibrated-QA" (C-QA) and "Write" in \textsc{Adapt}, with memory usages in "Clarify" (short-term) and "Write" (long-term).
}
\label{fig:model}
\end{figure*}


\subsection{Task Definition}

\textit{Exemplar-Based Expository Text Generation} task is formally defined as: Given an expository text $\mathbf{X}=\{x_i\}_{i=1}^m$ of a sequence of sentences on a source topic $\mathbf{t_x}$, a target topic $\mathbf{t_y}$, and external knowledge sources denoted as $\mathbf{K}$, the objective is to produce a new expository text $\mathbf{Y}=\{y_j\}_{j=1}^n$ on the target topic $\mathbf{t_y}$ that imitates the content and structure of $\mathbf{X}$ while adapting topic-specific content. 

Our goal is to develop an instructed imitative content creator, a system capable of generating a long text on a target topic given a provided long exemplar text on a source topic. This task, essentially knowledge-intensive one-shot long-form text generation, involves comprehending the structure and content of the source document, extracting pertinent information about the source topic, and then adapting this information to ensure relevance and accuracy for the target topic. Crucially, the challenge lies in preserving the coherence and quality of the source text while ensuring the accuracy and relevance of the generated output to the target topic. 

\subsection{\textsc{\textbf{RePA}: Recurrent Plan-then-Adapt}}\label{sec:recurrent}

We propose a novel \textsc{Recurrent Plan-then-Adapt} (\textsc{\textbf{RePA}}) model, outlined in Algorithm~\ref{alg:overall}. \textsc{RePA} first segments\footnote{We opt for sentences as the recurrence basis as they are semantic units that strike a balance between granularity and coherence for imitation and adaptation.} the input text $\mathbf{X}$ into a sequence of text segments $\{x_t\}_{t=1}^T$, then recurrently processes each input segment $x_t$ and generates an output segment $y_t$. At each recurrence step, \textsc{RePA} employs a \textsc{Plan-then-Adapt} process. The \textsc{Plan} stage involves learning outlines $q_t$ from input $x_t$ as plans for generation. The \textsc{Adapt} stage is to realize outlines $q_t$ flexibly and effectively with knowledge $\mathbf{K}$ to generate the output $y_t$ adaptively. Specifically, we introduce two on-the-fly memory structures to retain essential information from previous recurrent steps: a short-term memory $h_t$ for the history input segments and a long-term memory $c_t$ for the history output segments. This design ensures our model's capability to handle arbitrarily long text without sacrificing information.

\begin{algorithm}
\small
\caption{\textsc{Recurrent Plan-then-Adapt}}\label{alg:overall}
\begin{algorithmic}
\State \textbf{Initialize} $\mathbf{Y}$ as an empty string
\For{$x_t \in \{x_t\}_{t=1}^T \in \mathbf{X}$}
\State $q_t, h_{t+1} \gets \textsc{Plan}(x_t, h_t)$
\State $y_t, c_{t+1} \gets \textsc{Adapt}(q_t, c_t, \mathbf{K})$
\State Append $y_t$ to $\mathbf{Y}$
\EndFor \\
\Return $\mathbf{Y}$
\end{algorithmic}
\end{algorithm}


As showed in Figure~\ref{fig:model}, \textsc{RePA} mirrors the recurrence structure in LSTMs, but a closer comparison highlights several distinct features: 1) it employs a text-based representation for input $x_t$, output $y_t$, short-term memory $h_t$, and long-term memory $c_t$; 2) it leverages prompting of general-purpose LLMs $\mathcal{M}$ such as GPT-4 to perform computations; and 3) it integrates the LLM-driven \textsc{Plan-then-Adapt} process within each recurrent unit. Formally: 
\begin{equation*}
    y_t, c_{t+1}, h_{t+1} = \text{Recurrent}(x_t, c_t, h_t, \mathcal{M}, \mathbf{K}),
\end{equation*}
where $x_t$, $y_t$, $h_t$, $c_t$ denote input segment, output segment, shot-term memory, long-term memory at time step $t$, respectively. $\mathbf{K}$ denotes external knowledge. The final output is a concatenation of all output segments, i.e., $\mathbf{Y}=y_1 y_2 \cdots y_T$. Figure~\ref{fig:model} also presents a running example, with full examples available in Appendix~\ref{appendix:example}.

\subsubsection{The \textsc{Plan} Module}\label{sec:plan}

Content planning plays a crucial role in guiding long-form text generation. However, unlike traditional approaches that rely on external "bank of plans" for planning~\cite{hua-etal-2019-argument-generation, goldfarb-tarrant-etal-2020-content, balepur-etal-2023-expository}, our task necessitates deriving a content plan exclusively from a single lengthy input text. This constraint imposes strict requirements on maintaining fidelity to the original input's content and structure. Drawing inspiration from the "Questions under Discussion" (QUD) theory, recent works have conceptualized text plans as sequences of question-answer pairs~\cite{narayan-etal-2023-conditional, huot2023text} for query-focused summarization. However, such approaches often require exhaustive annotation for question generation~\cite{liu-etal-2023-ask}. To grasp the content and structure of the input text without plan annotation, we aim to extract key information and frame it into questions by prompting LLMs, which serve as outlines to guide subsequent generation, helping to retain \emph{cross-topic consistency}. Our \textsc{Plan} module comprises two sequential components: Clarify and Outline. 

\paragraph{Clarify}
Segmenting long inputs into smaller segments can introduce coreference ambiguities, where pronouns may lack clear antecedents in a current segment $x_t$, \textit{e.g.}, "it" in the input text segment (Figure~\ref{fig:model}) is unclear. This ambiguity complicates the understanding of the current input. To address this, we introduce the Clarify component, which uses short-term memory, $h_t$, to resolve coreference issues, e.g., replacing "it" with "Belebeyevsky District" for better comprehension. The memory $h_t$ stores key information from \textit{recent} input segments and is updated each time a clarified segment is generated. Essentially, this short-term memory acts as a sliding window of context, helping to \textit{clarify} the current segment by replacing pronouns with their corresponding antecedents (Figure~\ref{fig:prompt-clarify}). 
Formally: 
\begin{equation*}
    x_t', h_{t+1} = \text{LLM}_\text{Clarify}(x_t, h_t),
\end{equation*}

\paragraph{Outline}
The Outline component then frames key points into outlines, initially focusing on the source topic $\mathbf{t_x}$ before transferring to the target topic $\mathbf{t_y}$. The outlines are conceptualized as a set of topic-centric questions on topic $\mathbf{t_y}$, achieved by first generating questions from clarified input segment and replacing source topic tokens with the target ones in the questions. Essentially, the Outline component performs question generation and transfer (Figure~\ref{fig:prompt-outline}), as examples in Figure~\ref{fig:runningexample}. The advantage of using questions as outlines lies in their conciseness for summarizing key points and their transitivity for transferring key points from source topic to target topic with simply topic token substitution. Thus, we have:
\begin{equation*}
    q_t = \text{LLM}_\text{Outline}(x_t', t_x, t_y)
\end{equation*}

\subsubsection{The \textsc{Adapt} Module}\label{sec:adapt}

Previous \textit{plan-then-generate} frameworks often assume plans will seamlessly translate into effective outputs. However, in our task, plans for the target topic are derived from those for the source topic. Ideally, talking points would align perfectly between the source and target topics, but in reality, they may vary, making the outlines less suitable for the target topic, \emph{e.g.}, "Chuvash name" does not exist for target topic in Figure~\ref{fig:illustration}. Since our goal is to generate informative texts which maintain factual correctness, it is essential to \emph{adapt} the outlines for the target topic. Therefore, we propose an \textsc{Adapt} module to handle outlines flexibly and gracefully, accommodating \textit{cross-topic variability}. The intuition of this module is that "an imperfect outline for the target topic is acceptable if handled correctly". Specifically, there are two components: Calibrated Question Answering (Calibrated-QA) and Write. 

\paragraph{Calibrated-QA}
A straightforward approach to handling outlines in the form of questions is retrieval-augmented question answering (QA). However, outlines may not perfectly align with the target topic, which can result in unsuitable or unanswerable questions. To address this, we introduce a \textit{refusal} mechanism inspired by recent works on confidence calibration~\cite{xiong2023can, tian-etal-2023-just}, which employ verbalized confidence to calibrate black-box LLMs and have demonstrated success in verbalized confidence calibration for factoid short-answer QA. Our approach involves prompting LLMs to generate confidence in their answers (Figure~\ref{fig:prompt-calibratedQA}), and to refuse to answer questions deemed unanswerable by filtering out those with low-confidence answers, as an example in Figure~\ref{fig:illustration} where "NA" corresponds to a low-confidence answer. This method shifts the focus from generating a perfect outline to handling it flexibly and appropriately. For retrieval, we conduct both per-topic retrieval using the target topic and per-query retrieval using the current question. We apply in-context retrieval-augmented language models (RALM)~\cite{ram2023context} for calibrated QA. Essentially, Calibrated-QA involves retrieval-augmented QA with verbalized confidence calibration, denoted as:
\begin{align*}
    k_t &= \text{Retriever}(t_y, q_t), \\
    a_t &= \text{LLM}_\text{Calibrated-QA}(q_t, k_t,\theta),
\end{align*}
where $k_t$ is knowledge, and $a_t$ are answers with verbalized confidences higher than threshold $\theta$ (or adapted facts as shown in Figure~\ref{fig:model}). 

\paragraph{Write}
Based on the adapted facts, the Write component aims to generate an output segment which aligns with the content and structure of the input, maintains coherency and avoids repetition in output. It first generates a draft output segment that is consistent with the adapted facts on the target topic, then revises it to remove redundant content from previous output segments (Figure~\ref{fig:prompt-write}), with a complete example in Figure~\ref{fig:runningexample}. To support this process, we introduce a long-term memory $c_t$ to store key information from \textit{all} previous output segments, and $c_t$ is updated to summarize the latest output segment once the current target segment is generated. Formally:
\begin{align*}
    &y_t = \text{LLM}_\text{Write}(a_t, c_t), \\
    &c_{t+1} = \text{LLM}_\oplus (c_t, y_t),
\end{align*}
where $\oplus$ denotes summarization (Figure~\ref{fig:prompt-summarize}). 
\section{Experimental Setup}\label{sec:setup}

\subsection{Datasets}

To ensure effective imitation and adaptation, source and target topics must belong to the same domain. Such "correspondence" demands a high level of similarity, though "variations" usually exist. To this end, we collect three diverse datasets\footnote{Our datasets consist of long texts, though not as extensive as those spanning thousands of words as we prioritize high-quality, similar text pairs with acceptable variance; we also tested longer texts, as shown in Appendix~\ref{appendix:longer-texts}.} to cover both open-domain and domain-specific scenarios, including Wikipedia, RoleEE, and USNews. 


\paragraph{Wikipedia}
The Wikipedia dataset is domain-agnostic. We collected the overview sections of open-domain Wikipedia articles, with Wikipedia titles serving as topics. Specifically, we used the English Wikipedia dump as of April 1st, 2024\footnote{https://dumps.wikimedia.org/enwiki/20240401/}, and trained title embeddings using Wikipedia2Vec\footnote{https://wikipedia2vec.github.io/wikipedia2vec/}~\cite{yamada-etal-2020-wikipedia2vec}. We then employed cosine similarity to pair Wikipedia topics and texts, ensuring a similarity score higher than 0.95. To enhance text quality and similarity, we filtered out texts with fewer than 3 sentences or fewer than 60 words, as well as topic pairs with significant divergence in their categories\footnote{https://en.wikipedia.org/wiki/Wikipedia:Contents/Categories}, indicated by a percentage of common category tags higher than 0.3. We totally collected 1000 samples and a random manual check of the collected Wikipedia dataset showed its suitability for our task.

\paragraph{RoleEE}
The RoleEE dataset is a multi-domain event dataset introduced by \citet{jiao-etal-2022-open}, featuring 50 impactful hot event types. We selected three event categories--academic award ceremony, music award ceremony, and satellite launch--and took the corresponding events as topics. After manually removing poor-quality texts within each category, we paired them using the Hugging Face text embedding tool\footnote{https://huggingface.co/jinaai/jina-embeddings-v2-base-en} to obtain the top 500 topic/text pairs. Specifically, for each event, we retained the first paragraph from its text to ensure text similarity.

\paragraph{USNews}
The USNews dataset is a domain-specific dataset from U.S. News best colleges\footnote{https://www.usnews.com/best-colleges}. We crawled the overviews of 420 best national universities, with universities serving as topics. These were then paired based on cosine similarity using the same Hugging Face text embedding tool to obtain the top 500 topic/text pairs. Specifically, for each university, we extracted the first paragraph from its overview section, and the first sentence of the second paragraph if available, considering the high similarity of first paragraphs across universities and the significant divergence in subsequent paragraphs of their overviews. 



\subsection{Baselines}\label{sec:baselines}

We employ the following baselines, and LLMs denote GPT-4~\cite{achiam2023gpt} or LLaMA 3\footnote{https://LLaMA.meta.com/LLaMA3/}. 

\noindent \textbf{LLM} is to directly prompt an LLM to generate a target text given the source text and the target topic. 

\noindent \textbf{RollingLLM (RoM)} is to divide input text into segments and recurrently prompt an LLM to generate a target segment until reaching the last input segment, similar to using a sliding window approach for generating long texts with Transformers. 

\noindent \textbf{o1} is a new OpenAI model\footnote{https://openai.com/index/introducing-openai-o1-preview/} that excel at solving complex problems by spending more time thinking before responding.

\noindent \textbf{Self-Refine (SR)}~\cite{madaan2024self} is to improve initial outputs through iterative feedback and refinement. We divide the input text into segments and perform 4 iterations of Self-Refine on each segment for comparable inference steps with \textsc{RePA}.

\noindent \textbf{Default} is to simply replace source topic in the source text with target topic, serving as an approximation for understanding cross-topic variability. 

We also developed retrieval-augmented generation (RAG) variants of the aforementioned LLM-based baselines to integrate retrieved knowledge on the target topic during generation, specifically \textbf{LLM+Retr}, \textbf{RoM+Retr}, \textbf{o1+Retr}, and \textbf{SR+Retr}.





\subsection{Implementation Details}


\textsc{RePA} is built upon two latest general-purpose Language Models (LLMs) in our experiments: OpenAI GPT-4~\cite{achiam2023gpt} and Meta LLaMA 3\footnote{https://LLaMA.meta.com/LLaMA3/}. Specifically, we employed the models "gpt-4-0125-preview", "meta/meta-LLaMA-3-8b-instruct", and "meta/meta-LLaMA-3-70b-instruct". o1 is actually the OpenAI model "o1-preview-2024-09-12". For these LLMs, we configure the following parameters: temperature as 0.3, frequency penalty as 0.3, max generation tokens as 256, and the number of response choices as 1. Additionally, for calibrated QA, we set the confidence threshold as 0.7, and aimed to make the process as deterministic as possible by setting the seed. The prompts used in each module are provided in Appendix~\ref{appendix:prompts-model}. 

For retrieval, we employed both per-topic and per-query retrieval. For topic retrieval, we utilized the Bing API as the retriever, with the topics as the input. This applies to both our model and the concerned baselines. Given that datasets derived from Wikipedia articles require knowledge from the open web, we excluded Wikipedia domains when using the Bing API\footnote{Wikipedia domains: "wikipedia.org", "wikiwand.com", "wiki2.org", and "wikimedia.org".}. Similarly, for datasets sourced from U.S. News, we excluded U.S. News domains\footnote{U.S. News domains: "usnews.com"}. For query retrieval, we employed bi-encoder in DPR~\cite{karpukhin-etal-2020-dense} as the retriever to retrieve from a fixed knowledge base, comprising sentence-level facts. Specifically, we used the "multiset" versions of question and context encoders, respectively, and used their cosine similarity for retrieval. When incorporating knowledge pieces into the prompts, we selected the top 10 Bing results and the top 3 DPR results.

\begin{table*}[t!]
  \footnotesize
  \centering
  \scalebox{0.92}{\begin{tabular}{llllllllllll}
  \toprule
    \textbf{Datasets} & \textbf{Models} & \textbf{R1$\uparrow$} & \textbf{R2$\uparrow$} & \textbf{R$\uparrow$L} & \textbf{RLsum$\uparrow$} & \textbf{Meteor$\uparrow$} & \textbf{BLEU$\uparrow$} & \textbf{Halluc$\downarrow$} & \textbf{NLI-E$\uparrow$} & \textbf{NLI-C$\downarrow$} \\
    \midrule
    \multirow{9}*{Wikipedia}
    & \textsc{RePA} & \textbf{0.8112} & \textbf{0.7146} & \textbf{0.7600} & \textbf{0.7625} & 0.7368 & \textbf{0.6672} & \textbf{6.5714} & \textbf{0.7927} & \textbf{0.0439} \\
    & LLM & 0.6855 & 0.4835 & 0.6154 & 0.6133 & 0.6984 & 0.4236 & 23.5794 & 0.3604 & 0.4561 \\
    & LLM+Retr & 0.7583 & 0.6121 & 0.7027 & 0.7057 & 0.7595 & 0.5887 & 15.8273 & 0.5343 & 0.3106 \\
    & RoM & 0.7300 & 0.5563 & 0.6785 & 0.6774 & 0.7448 & 0.5190 & 17.3810 & 0.3640 & 0.5223 \\
    & RoM+Retr & 0.7343 & 0.6073 & 0.6889 & 0.6906 & 0.7328 & 0.5408 & 14.9102 & 0.5240 & 0.3536 \\
    & o1 & 0.7658 & 0.6345 & 0.7223 & 0.7210 & 0.7547 & 0.6018 & 14.4050 & 0.4954 & 0.3720 \\
    & o1+Retr & 0.7421 & 0.6258 & 0.6993 & 0.7002 & 0.6646 & 0.5804 & 13.2928 & 0.6227 & 0.2194 \\
    & SR & 0.7387 & 0.5628 & 0.6782 & 0.6748 & 0.6957 & 0.5225 & 21.6280 & 0.3701 & 0.4653 \\
    & SR+Retr & 0.7839 & 0.6247 & 0.7187 & 0.7202 & 0.7599 & 0.6011 & 10.6257 & 0.6032 & 0.2751 \\
    & Default & 0.7520 & 0.6042 & 0.7129 & 0.7127 & \textbf{0.7667} & 0.5705 & 16.4128 & 0.1311 & 0.7152 \\
    \midrule
    \multirow{9}*{RoleEE}
    & \textsc{RePA} & \textbf{0.9184} & \textbf{0.8691} & \textbf{0.9029} & \textbf{0.9027} & \textbf{0.9170} & \textbf{0.7548} & \textbf{5.2604} & \textbf{0.9067} & \textbf{0.0653} \\
    & LLM & 0.6780 & 0.5329 & 0.6399 & 0.6526 & 0.7200 & 0.3784 & 26.9641 & 0.2696 & 0.5487 \\
    & LLM+Retr & 0.8848 & 0.8197 & 0.8692 & 0.8726 & 0.9093 & 0.6924 & 9.4551 & 0.7109 & 0.1833 \\
    & RoM & 0.7910 & 0.6572 & 0.7735 & 0.7747 & 0.8094 & 0.6096 & 15.3346 & 0.2543 & 0.7003 \\
    & RoM+Retr & 0.8842 & 0.8191 & 0.8715 & 0.8715 & 0.8954 & 0.7117 & 6.4164 & 0.7400 & 0.2053 \\
    & o1 & 0.8306 & 0.7273 & 0.8165 & 0.8136 & 0.8334 & 0.6932 & 13.1344 & 0.3793 & 0.5457 \\
    & o1+Retr & 0.8849 & 0.8233 & 0.8706 & 0.8671 & 0.8812 & 0.7814 & 9.0228 & 0.7583 & 0.1673 \\
    & SR & 0.8126 & 0.7856 & 0.7892 & 0.7821 & 0.8255 & 0.6241 & 18.0331 & 0.2527 & 0.6415 \\
    & SR+Retr & 0.9041 & 0.8306 & 0.8652 & 0.8624 & 0.9083 & 0.7262 & 7.7313 & 0.7981 & 0.1376 \\
    & Default & 0.8090 & 0.6722 & 0.7898 & 0.7893 & 0.8351 & 0.6003 & 15.7650 & 0.1200 & 0.8301 \\
    \midrule
    \multirow{9}*{USNews}
    & \textsc{RePA} & \textbf{0.8922} & \textbf{0.8396} & \textbf{0.8642} & \textbf{0.8653} & \textbf{0.8971} & \textbf{0.8129} & \textbf{4.6048} & \textbf{0.8085} & \textbf{0.0441} \\
    & LLM & 0.7651 & 0.6258 & 0.7258 & 0.7270 & 0.8043 & 0.5427 & 18.3919 & 0.3749 & 0.5257 \\
    & LLM+Retr & 0.5842 & 0.4303 & 0.5234 & 0.5249 & 0.5994 & 0.3037 & 33.9516 & 0.3836 & 0.2853 \\
    & RoM & 0.7390 & 0.6178 & 0.7156 & 0.7153 & 0.8008 & 0.5585 & 15.7913 & 0.3621 & 0.4873 \\
    & RoM+Retr & 0.6561 & 0.5127 & 0.6150 & 0.6172 & 0.6866 & 0.4312 & 23.4262 & 0.3322 & 0.4047 \\
    & o1 & 0.8327 & 0.7404 & 0.8138 & 0.8144 & 0.8548 & 0.6970 & 12.1111 & 0.4200 & 0.4827 \\
    & o1+Retr & 0.7265 & 0.6082 & 0.6880 & 0.6875 & 0.7273 & 0.5572 & 19.4480 & 0.3970 & 0.3372 \\
    & SR & 0.7381 & 0.6137 & 0.7019 & 0.6973 & 0.8126 & 0.5495 & 19.5238 & 0.3527 & 0.4562 \\
    & SR+Retr & 0.8458 & 0.7239 & 0.8182 & 0.8139 & 0.8639 & 0.7236 & 11.5269 & 0.5338 & 0.1757 \\
    & Default & 0.7245 & 0.5808 & 0.6999 & 0.7013 & 0.7768 & 0.5494 & 18.1090 & 0.1667 & 0.5700 \\
    \bottomrule
  \end{tabular}}
  \vspace{-0.05in}
  \caption{Evaluation results on basic and factuality metrics. LLM denotes GPT-4 specifically.}
  \label{tab:comparison-1-gpt}
  \vspace{-0.15in}
\end{table*}

\subsection{Metrics}\label{sec:metrics}

\noindent \textbf{Basic Metrics}\quad
To provide a basic assessment of this generation task, we use well-established metrics commonly used in previous research, namely ROUGE~\cite{lin-2004-rouge}, BLEU~\cite{papineni-etal-2002-bleu}, and METEOR~\cite{banerjee-lavie-2005-meteor}. 

\noindent \textbf{Task-Specific Metrics}\quad
Established metrics cannot capture the nuances of our task. We leverage LLM-as-a-Judge~\cite{zheng2024judging} to evaluate \emph{Imitativeness} and \emph{Adaptiveness}. Imitativeness represents the model's ability to faithfully replicate the structure and content of the exemplar, ensuring cross-topic consistency. Adaptiveness assesses how effectively the model adapts the exemplar content to a new topic, addressing cross-topic variability while ensuring consistency. We also introduce \textit{Adaptive-Imitativeness}, which is a F1-score of Imitativeness and Adaptiveness. Additional details and discussion on addressing the known limitations of LLM-Judge are provided in Appendix~\ref{appendix:LLMmetrics}. 

\noindent \textbf{Factuality Metrics}\quad
Given the knowledge-intensive nature of our task, maintaining factual accuracy is crucial. Inspired by FActScore~\cite{min-etal-2023-factscore}, a metric for evaluating factual accuracy of long-form text, we decompose output text into sentence-level facts, then take ground truth as knowledge source to calculate the percentages of entailment (\textbf{NLI-E}) and contradiction (\textbf{NLI-C}) sentences in outputs. Our human evaluation of the effectiveness of NLI-based metrics are included in Appendix~\ref{appendix:NLImetrics-eval}. We also compute the percentage of hallucinated tokens compared to both inputs and ground truths, named \textbf{Halluc}. 
\section{Experimental Results}\label{sec:results}

\subsection{Comparison with Baselines}

\begin{table}[t!]
  \footnotesize
  \centering
  \scalebox{0.92}{\begin{tabular}{lllll}
    \toprule
    \textbf{Datasets} & \textbf{Models} & \textbf{\textit{I.}$\uparrow$} & \textbf{\textit{A.}$\uparrow$} & \textbf{\textit{A.-I.}$\uparrow$} \\
    \midrule
    \multirow{9}*{Wikipedia}
    & \textsc{RePA} & 4.16 & \textbf{3.90} & \textbf{3.93} \\
    & LLM & 4.52 & 2.44 & 3.06 \\
    & LLM+Retr & 4.46 & 2.78 & 3.25 \\
    & RoM & \underline{4.58} & 2.32 & 3.00 \\
    & RoM+Retr & 4.08 & 2.56 & 3.00 \\
    & o1 & 4.34 & 2.94 & 3.40 \\
    & o1+Retr & 4.32 & 3.02 & 3.42 \\
    & SR & 4.56 & 2.54 & 3.25 \\
    & SR+Retr & 4.22 & \underline{3.04} & \underline{3.50} \\
    & Default & \textbf{5.00} & 1.08 & 1.73 \\
    \midrule
    \multirow{9}*{RoleEE}
    & \textsc{RePA} & \underline{4.80} & \textbf{4.30} & \textbf{4.46} \\
    & LLM & 4.70 & 2.76 & 3.23 \\
    & LLM+Retr & \underline{4.80} & \underline{4.26} & \underline{4.39} \\
    & RoM & 4.62 & 1.94 & 2.55 \\
    & RoM+Retr & 4.70 & 4.04 & 4.21 \\
    & o1 & 4.64 & 2.66 & 3.12 \\
    & o1+Retr & 4.68 & 4.24 & 4.34 \\
    & SR & 4.70 & 2.62 & 3.30 \\
    & SR+Retr & 4.74 & 4.22 & 3.33 \\
    & Default & \textbf{5.00} & 1.24 & 1.87 \\
    \midrule
    \multirow{9}*{USNews}
    & \textsc{RePA} & 4.22 & \textbf{4.32} & \textbf{4.22} \\
    & LLM & 4.20 & 3.06 & 3.45 \\
    & LLM+Retr & 4.02 & 2.86 & 3.25 \\
    & RoM & \underline{4.58} & 2.74 & 3.30 \\
    & RoM+Retr & 4.08 & 2.40 & 2.96 \\
    & o1 & 4.20 & 3.18 & 3.53 \\
    & o1+Retr & 4.14 & 2.98 & 3.37 \\
    & SR & 4.44 & 2.98 & 3.49 \\
    & SR+Retr & 4.12 & \underline{3.74} & \underline{3.84} \\
    & Default & \textbf{5.00} & 1.00 & 1.67 \\
    \bottomrule
  \end{tabular}}
  \vspace{-0.05in}
  \caption{Evaluation results on task-specific metrics. LLM denotes GPT-4 specifically. \textit{\textbf{I.}} denotes \textit{Imitativeness}, \textbf{\textit{A.}} denotes \textit{Adaptiveness}, and \textbf{\textit{A.-I.}} denotes \textit{Adaptive-Imitativeness}.}
  \label{tab:comparison-2-gpt}
  \vspace{-0.15in}
\end{table}

We present the evaluation results of models built on GPT-4 in Tables~\ref{tab:comparison-1-gpt} and \ref{tab:comparison-2-gpt}. Results of models built on LLaMA 3 are included in Appendix~\ref{appendix:llama-results}. We also conducted a case study as in Appendix~\ref{appendix:case-study}. 

\begin{table*}[hbt!]
  \footnotesize
  \centering
  \scalebox{0.93}{\begin{tabular}{rlllllllllllll}
  \toprule
    & \textbf{R1$\uparrow$} & \textbf{R2$\uparrow$} & \textbf{RL$\uparrow$} & \textbf{RLsum$\uparrow$} & \textbf{Meteor$\uparrow$} & \textbf{BLEU$\uparrow$} & \textbf{Halluc$\downarrow$} & \textbf{NLI-E$\uparrow$} & \textbf{NLI-C$\downarrow$} & \textbf{\textit{I.}$\uparrow$} & \textbf{\textit{A.}$\uparrow$} & \textbf{\textit{A.-I.}$\uparrow$} \\
    \midrule
    Full & \textbf{0.8112} & \textbf{0.7146} & \textbf{0.7600} & \textbf{0.7625} & \textbf{0.7368} & \textbf{0.6672} & \textbf{6.5714} & \textbf{0.7927} & \textbf{0.0439} & \textbf{4.16} & \textbf{3.90} & \textbf{3.93} \\
    - C & 0.8000 & 0.6958 & 0.7369 & 0.7395 & 0.7149 & 0.6341 & 6.7293 & 0.7826 & 0.0539 & \textbf{4.16} & 3.72 & 3.83 \\
    - O & 0.5879 & 0.4574 & 0.5215 & 0.5218 & 0.5928 & 0.3415 & 25.3955 & 0.5847 & 0.0850 & 3.82 & 2.88 & 3.15 \\
    - F & 0.8039 & 0.7121 & 0.7581 & 0.7571 & 0.7204 & 0.6543 & 7.0788 & 0.7481 & 0.0859 & 4.08 & 3.54 & 3.69 \\
    - R & 0.7775 & 0.6634 & 0.7154 & 0.7167 & 0.7065 & 0.6037 & 6.5915 & 0.7898 & 0.0510 & 4.10 & 3.74 & 3.84 \\
    - S & 0.7482 & 0.6365 & 0.6627 & 0.6679 & 0.6562 & 0.5770 & 8.2648 & 0.7135 & 0.0683 & 4.06 & 3.43 & 3.74 \\
    \bottomrule
  \end{tabular}}
  \vspace{-0.05in}
  \caption{Evaluation results across all metrics for ablation study on Wikipedia dataset. LLM denotes GPT-4 specifically. - C, - O, - F, - R, -S denote model variants w/o Clarify-STM, w/o Outline, w/o Refusal, w/o Revise-LTM, w/o Segment, respectively.}
  \label{tab:ablation-gpt}
  \vspace{-0.15in}
\end{table*}

Considering basic generation metrics, \textsc{RePA} outperforms baselines on almost all metrics, although LLaMA 3-based baselines may achieve higher scores on metrics such as ROUGE and Meteor. Notably, the Default baseline, which simply replaces the source topic with the target topic in the source text, achieves a Meteor score of 0.7667 on the Wikipedia dataset, higher than all GPT-4 based models, suggesting that \textbf{established basic metrics might not be a reliable indicator of model performance for our task}. For factuality metrics including Halluc, NLI-E and NLI-C, \textsc{RePA} significantly outperforms baselines including retrieval-augmented baselines, indicating that \textbf{\textsc{RePA} generates more factual content}. 


For task-specific metrics \textit{Imitativeness}, \textit{Adaptiveness}, and \textit{Adaptive-Imitativeness}, we find that prompting LLMs yields high Imitativeness, showing \textbf{LLMs are strong imitators regardless of output factuality}. However, their extremely low Adaptiveness indicates poor cross-topic variability recognition and weak topic-specific adaptation. Moreover, the Default baseline achieves the highest Imitativeness, suggesting that \textbf{Imitativeness alone is insufficient} and Adaptive-Imitativeness is needed for comprehensive model evaluation. In contrast, our proposed \textsc{RePA} prioritizes Adaptiveness and Adaptive-Imitativeness over Imitativeness, striking a balance that leads to the best overall performance for this task. Its task-specific design enables superior adaptive imitation and ensures consistent results across diverse datasets.


Additionally, our collected datasets have varying degrees of correspondence and variation. The RoleEE dataset reflects high correspondence (low variation), USNews medium, and Wikipedia relatively low (high variation). Results in Table~\ref{tab:comparison-1-gpt} and \ref{tab:comparison-2-gpt} show better model performance with higher correspondence and lower variation. Overall, the results highlight \textsc{RePA}'s superior ability to generate factual texts with high imitativeness and adaptiveness.

\subsection{Ablation Study}

To study how different components of \textsc{RePA} contribute to its overall performance, we conduct an ablation study with the following variations:

\noindent \textbf{1) w/o Clarify-STM} removes the Clarify component in \textsc{Plan}, as well as the \textit{short-term memory} for retaining history processed input segments. 

\noindent \textbf{2) w/o Outline} removes the Outline component in \textsc{Plan}. Consequently, Calibrated-QA is also removed, and retrieval is based solely on the clarified input segment without any generated questions. 
    
\noindent \textbf{3) w/o Refusal} removes confidence calibration in QA, resulting in no refusal in answering questions. 

\noindent \textbf{4) w/o Revise-LTM} removes the Revise component in Write, as well as the \textit{long-term memory} for storing history generated output.

\noindent \textbf{5) w/o Segment} removes text segmentation, and the input exemplar text was processed as a single block in the \textsc{Plan-then-Adapt} process, bypassing recurrent steps. 

We evaluate these variants with the Wikipedia dataset on all metrics. As the GPT 4-based results shown in Table~\ref{tab:ablation-gpt} (additional results on LLaMA 3-based variants are included in Appendix~\ref{appendix:llama-results}), we find that the Outline (or \textsc{Plan}) component significantly contributes to the overall model performance, demonstrating the effectiveness of using questions as a format of outlines for guiding factual adaptive-imitative generation. Additionally, the Refusal mechanism enhances the generation of factual content; removing it results in a decrease in factuality. Overall, the full \textsc{RePA} model exhibits the best performance across all metrics on the Wikipedia dataset, and each component of \textsc{RePA} plays a crucial role in improving its overall performance.

\subsection{Human Evaluation of LLM Judge}

Following the methodology of \citet{zheng2024judging}, we measured the agreement between the LLM-judge and human annotators, calculated as the probability that both parties would select the same model output from a randomly chosen pair. As shown in Table~\ref{tab:human-brief}, there is \textbf{a strong correlation between LLM and human judgments}, with mean agreement scores of 79.0\% for Imitativeness and 82.9\% for Adaptiveness. These results demonstrate that our LLM-judge metrics align closely with human judgments, validating their reliability. Moreover, the LLM-judge's agreement rates were comparable to or exceeded inter-annotator agreement among humans, which was 79.2\% for Imitativeness and 80.8\% for Adaptiveness. Additional details are included in Appendix~\ref{appendix:LLMmetrics-eval}. 

\vspace{-0.1in}
\begin{table}[hbt!]
  \footnotesize
  \centering
  \begin{tabular}{lll}
  \toprule
    \textbf{Dataset} & \textbf{\textit{I.} - w/o tie} & \textbf{\textit{A.} - w/o tie} \\
    \midrule
    LLM-Human & 79.0\% & 82.9\% \\
    Human-Human & 79.2\% & 80.8\% \\
    \bottomrule
  \end{tabular}
  \vspace{-0.05in}
  \caption{Agreements on Imitativeness (\textit{I.}) and Adaptiveness (\textit{A.}) metrics.}
  \label{tab:human-brief}
  \vspace{-0.15in}
\end{table}
\section{Conclusion and Discussion}\label{sec:conclusion}

In summary, we introduce a new, practical yet under-explored task: \textit{Exemplar-Based Expository Text Generation}. To ensure \textit{cross-topic consistency}, address \textit{cross-topic variability}, and scale to long-form text, we propose \textsc{\textbf{RePA}} (\textsc{Recurrent Plan-then-Adapt}), a model incorporating two memory structures--a short-term memory and a long-term memory. To address the limitations of existing evaluation metrics, we employ LLM-as-a-Judge to develop task-specific evaluators alongside established metrics. Extensive comparisons and ablation studies on three diverse, newly collected datasets demonstrate the effectiveness of our model across basic generation metrics, factuality metrics, and task-specific metrics, including \textit{imitativeness}, \textit{adaptiveness}, and \textit{adaptive-imitativeness}.

\section*{Limitations}

While our \textsc{RePA} model demonstrates promising results, several limitations remain. 
Firstly, our proposed \textit{Exemplar-Based Expository Text Generation} task is explicitly designed to generate expository texts using \textbf{a high-quality exemplar from a similar topic}. The goal of our task is to "\textit{write like the best}", which inherently assumes that a well-crafted and topically relevant exemplar is provided. The quality and topical alignment of the exemplar are fundamental premises of our approach, and our study is scoped around these assumptions. However, scenarios involving low-quality or dissimilar exemplars pose challenges to the model's performance, and future work might explore these scenarios. Additionally, although our recurrent prompting pipeline enhances performance, it may introduce inefficiencies compared to certain baseline methods. Moreover, the reliance on large language models (LLMs) such as GPT-4 and LLaMA 3 limits applicability in resource-constrained environments.

Future work could address these limitations in several ways. First, exploring methods for on-demand retrieval of relevant knowledge could mitigate the occasional inaccuracies generated by LLMs and improve overall efficiency. Second, incorporating multiple exemplars, rather than relying on a single one, may enhance generalization by broadening the diversity and scope of the output. Using exemplars from related topics could provide a more comprehensive perspective, enabling richer and more varied content generation.

\section*{Acknowledgments}
This material is based upon work supported by the National Science Foundation IIS 16-19302 and IIS 16-33755, Zhejiang University ZJU Research 083650, IBM-Illinois Center for Cognitive Computing Systems Research (C3SR) and IBM-Illinois Discovery Accelerator Institute (IIDAI), grants from eBay and Microsoft Azure, UIUC OVCR CCIL Planning Grant 434S34, UIUC CSBS Small Grant 434C8U, and UIUC New Frontiers Initiative. Any opinions, findings, conclusions, or recommendations expressed in this publication are those of the author(s) and do not necessarily reflect the views of the funding agencies.

\bibliography{anthology,custom}

\begin{thebibliography}{66}
\providecommand{\natexlab}[1]{#1}

\bibitem[{Achiam et~al.(2023)Achiam, Adler, Agarwal, Ahmad, Akkaya, Aleman,
  Almeida, Altenschmidt, Altman, Anadkat et~al.}]{achiam2023gpt}
Josh Achiam, Steven Adler, Sandhini Agarwal, Lama Ahmad, Ilge Akkaya,
  Florencia~Leoni Aleman, Diogo Almeida, Janko Altenschmidt, Sam Altman,
  Shyamal Anadkat, et~al. 2023.
\newblock Gpt-4 technical report.
\newblock \emph{arXiv preprint arXiv:2303.08774}.

\bibitem[{Adewoyin et~al.(2022)Adewoyin, Dutta, and
  He}]{adewoyin-etal-2022-rstgen}
Rilwan Adewoyin, Ritabrata Dutta, and Yulan He. 2022.
\newblock \href {https://doi.org/10.18653/v1/2022.naacl-main.133} {{RSTG}en:
  Imbuing fine-grained interpretable control into long-{F}orm{T}ext
  generators}.
\newblock In \emph{Proceedings of the 2022 Conference of the North American
  Chapter of the Association for Computational Linguistics: Human Language
  Technologies}, pages 1822--1835, Seattle, United States. Association for
  Computational Linguistics.

\bibitem[{Bai et~al.(2021)Bai, Li, Ding, Shen, and Zheng}]{bai2021infobox}
Yang Bai, Ziran Li, Ning Ding, Ying Shen, and Hai-Tao Zheng. 2021.
\newblock Infobox-to-text generation with tree-like planning based attention
  network.
\newblock In \emph{Proceedings of the Twenty-Ninth International Conference on
  International Joint Conferences on Artificial Intelligence}, pages
  3773--3779.

\bibitem[{Balepur et~al.(2023)Balepur, Huang, and
  Chang}]{balepur-etal-2023-expository}
Nishant Balepur, Jie Huang, and Kevin Chang. 2023.
\newblock \href {https://doi.org/10.18653/v1/2023.emnlp-main.729} {Expository
  text generation: Imitate, retrieve, paraphrase}.
\newblock In \emph{Proceedings of the 2023 Conference on Empirical Methods in
  Natural Language Processing}, pages 11896--11919, Singapore. Association for
  Computational Linguistics.

\bibitem[{Banerjee and Lavie(2005)}]{banerjee-lavie-2005-meteor}
Satanjeev Banerjee and Alon Lavie. 2005.
\newblock \href {https://aclanthology.org/W05-0909} {{METEOR}: An automatic
  metric for {MT} evaluation with improved correlation with human judgments}.
\newblock In \emph{Proceedings of the {ACL} Workshop on Intrinsic and Extrinsic
  Evaluation Measures for Machine Translation and/or Summarization}, pages
  65--72, Ann Arbor, Michigan. Association for Computational Linguistics.

\bibitem[{Beltagy et~al.(2020)Beltagy, Peters, and
  Cohan}]{beltagy2020longformer}
Iz~Beltagy, Matthew~E Peters, and Arman Cohan. 2020.
\newblock Longformer: The long-document transformer.
\newblock \emph{arXiv preprint arXiv:2004.05150}.

\bibitem[{Bowman et~al.(2015)Bowman, Angeli, Potts, and
  Manning}]{bowman-etal-2015-large}
Samuel~R. Bowman, Gabor Angeli, Christopher Potts, and Christopher~D. Manning.
  2015.
\newblock \href {https://doi.org/10.18653/v1/D15-1075} {A large annotated
  corpus for learning natural language inference}.
\newblock In \emph{Proceedings of the 2015 Conference on Empirical Methods in
  Natural Language Processing}, pages 632--642, Lisbon, Portugal. Association
  for Computational Linguistics.

\bibitem[{Carter et~al.(2018)Carter, Salamonson, Ramjan, and
  Halcomb}]{carter2018students}
Rebekah Carter, Yenna Salamonson, Lucie~M Ramjan, and Elizabeth Halcomb. 2018.
\newblock Students use of exemplars to support academic writing in higher
  education: An integrative review.
\newblock \emph{Nurse education today}, 65:87--93.

\bibitem[{Chen(2024)}]{chen2024exploring}
Jun Chen. 2024.
\newblock Exploring imitative learning in a blended efl writing class.

\bibitem[{Gatt and Krahmer(2018)}]{gatt2018survey}
Albert Gatt and Emiel Krahmer. 2018.
\newblock Survey of the state of the art in natural language generation: Core
  tasks, applications and evaluation.
\newblock \emph{Journal of Artificial Intelligence Research}, 61:65--170.

\bibitem[{Goldfarb-Tarrant et~al.(2020)Goldfarb-Tarrant, Chakrabarty,
  Weischedel, and Peng}]{goldfarb-tarrant-etal-2020-content}
Seraphina Goldfarb-Tarrant, Tuhin Chakrabarty, Ralph Weischedel, and Nanyun
  Peng. 2020.
\newblock \href {https://doi.org/10.18653/v1/2020.emnlp-main.351} {Content
  planning for neural story generation with aristotelian rescoring}.
\newblock In \emph{Proceedings of the 2020 Conference on Empirical Methods in
  Natural Language Processing (EMNLP)}, pages 4319--4338, Online. Association
  for Computational Linguistics.

\bibitem[{Guan et~al.(2020)Guan, Huang, Zhao, Zhu, and
  Huang}]{guan-etal-2020-knowledge}
Jian Guan, Fei Huang, Zhihao Zhao, Xiaoyan Zhu, and Minlie Huang. 2020.
\newblock \href {https://doi.org/10.1162/tacl_a_00302} {A knowledge-enhanced
  pretraining model for commonsense story generation}.
\newblock \emph{Transactions of the Association for Computational Linguistics},
  8:93--108.

\bibitem[{Guan et~al.(2021)Guan, Mao, Fan, Liu, Ding, and
  Huang}]{guan-etal-2021-long}
Jian Guan, Xiaoxi Mao, Changjie Fan, Zitao Liu, Wenbiao Ding, and Minlie Huang.
  2021.
\newblock \href {https://doi.org/10.18653/v1/2021.acl-long.499} {Long text
  generation by modeling sentence-level and discourse-level coherence}.
\newblock In \emph{Proceedings of the 59th Annual Meeting of the Association
  for Computational Linguistics and the 11th International Joint Conference on
  Natural Language Processing (Volume 1: Long Papers)}, pages 6379--6393,
  Online. Association for Computational Linguistics.

\bibitem[{Hu et~al.(2022)Hu, Chan, Liu, Xiao, Wu, and
  Huang}]{hu-etal-2022-planet}
Zhe Hu, Hou~Pong Chan, Jiachen Liu, Xinyan Xiao, Hua Wu, and Lifu Huang. 2022.
\newblock \href {https://doi.org/10.18653/v1/2022.acl-long.163} {{PLANET}:
  Dynamic content planning in autoregressive transformers for long-form text
  generation}.
\newblock In \emph{Proceedings of the 60th Annual Meeting of the Association
  for Computational Linguistics (Volume 1: Long Papers)}, pages 2288--2305,
  Dublin, Ireland. Association for Computational Linguistics.

\bibitem[{Hua et~al.(2019)Hua, Hu, and
  Wang}]{hua-etal-2019-argument-generation}
Xinyu Hua, Zhe Hu, and Lu~Wang. 2019.
\newblock \href {https://doi.org/10.18653/v1/P19-1255} {Argument generation
  with retrieval, planning, and realization}.
\newblock In \emph{Proceedings of the 57th Annual Meeting of the Association
  for Computational Linguistics}, pages 2661--2672, Florence, Italy.
  Association for Computational Linguistics.

\bibitem[{Hua et~al.(2023)Hua, Deng, and McKeown}]{hua-etal-2023-improving}
Yilun Hua, Zhaoyuan Deng, and Kathleen McKeown. 2023.
\newblock \href {https://doi.org/10.18653/v1/2023.findings-acl.871} {Improving
  long dialogue summarization with semantic graph representation}.
\newblock In \emph{Findings of the Association for Computational Linguistics:
  ACL 2023}, pages 13851--13883, Toronto, Canada. Association for Computational
  Linguistics.

\bibitem[{Huot et~al.(2023)Huot, Maynez, Narayan, Amplayo, Ganchev, Louis,
  Sandholm, Das, and Lapata}]{huot2023text}
Fantine Huot, Joshua Maynez, Shashi Narayan, Reinald~Kim Amplayo, Kuzman
  Ganchev, Annie~Priyadarshini Louis, Anders Sandholm, Dipanjan Das, and
  Mirella Lapata. 2023.
\newblock Text-blueprint: An interactive platform for plan-based conditional
  generation.
\newblock In \emph{Proceedings of the 17th Conference of the European Chapter
  of the Association for Computational Linguistics: System Demonstrations},
  pages 105--116.

\bibitem[{Ivgi et~al.(2023)Ivgi, Shaham, and Berant}]{ivgi-etal-2023-efficient}
Maor Ivgi, Uri Shaham, and Jonathan Berant. 2023.
\newblock \href {https://doi.org/10.1162/tacl_a_00547} {Efficient long-text
  understanding with short-text models}.
\newblock \emph{Transactions of the Association for Computational Linguistics},
  11:284--299.

\bibitem[{Ji et~al.(2023)Ji, Lee, Frieske, Yu, Su, Xu, Ishii, Bang, Madotto,
  and Fung}]{ji2023survey}
Ziwei Ji, Nayeon Lee, Rita Frieske, Tiezheng Yu, Dan Su, Yan Xu, Etsuko Ishii,
  Ye~Jin Bang, Andrea Madotto, and Pascale Fung. 2023.
\newblock Survey of hallucination in natural language generation.
\newblock \emph{ACM Computing Surveys}, 55(12):1--38.

\bibitem[{Jiang et~al.(2024)Jiang, Shao, Ma, Semnani, and
  Lam}]{jiang-etal-2024-unknown}
Yucheng Jiang, Yijia Shao, Dekun Ma, Sina Semnani, and Monica Lam. 2024.
\newblock \href {https://doi.org/10.18653/v1/2024.emnlp-main.554} {Into the
  unknown unknowns: Engaged human learning through participation in language
  model agent conversations}.
\newblock In \emph{Proceedings of the 2024 Conference on Empirical Methods in
  Natural Language Processing}, pages 9917--9955, Miami, Florida, USA.
  Association for Computational Linguistics.

\bibitem[{Jiang et~al.(2021)Jiang, Araki, Ding, and
  Neubig}]{jiang-etal-2021-know}
Zhengbao Jiang, Jun Araki, Haibo Ding, and Graham Neubig. 2021.
\newblock \href {https://doi.org/10.1162/tacl_a_00407} {How can we know when
  language models know? on the calibration of language models for question
  answering}.
\newblock \emph{Transactions of the Association for Computational Linguistics},
  9:962--977.

\bibitem[{Jiao et~al.(2022)Jiao, Li, Xie, Zhong, Ji, and
  Han}]{jiao-etal-2022-open}
Yizhu Jiao, Sha Li, Yiqing Xie, Ming Zhong, Heng Ji, and Jiawei Han. 2022.
\newblock \href {https://doi.org/10.18653/v1/2022.findings-emnlp.395}
  {Open-vocabulary argument role prediction for event extraction}.
\newblock In \emph{Findings of the Association for Computational Linguistics:
  EMNLP 2022}, pages 5404--5418, Abu Dhabi, United Arab Emirates. Association
  for Computational Linguistics.

\bibitem[{Jin et~al.(2024)Jin, Han, Yang, Jiang, Liu, Chang, Chen, and
  Hu}]{jin2024llm}
Hongye Jin, Xiaotian Han, Jingfeng Yang, Zhimeng Jiang, Zirui Liu, Chia-Yuan
  Chang, Huiyuan Chen, and Xia Hu. 2024.
\newblock Llm maybe longlm: Self-extend llm context window without tuning.
\newblock \emph{arXiv preprint arXiv:2401.01325}.

\bibitem[{Kadavath et~al.(2022)Kadavath, Conerly, Askell, Henighan, Drain,
  Perez, Schiefer, Hatfield-Dodds, DasSarma, Tran-Johnson
  et~al.}]{kadavath2022language}
Saurav Kadavath, Tom Conerly, Amanda Askell, Tom Henighan, Dawn Drain, Ethan
  Perez, Nicholas Schiefer, Zac Hatfield-Dodds, Nova DasSarma, Eli
  Tran-Johnson, et~al. 2022.
\newblock Language models (mostly) know what they know.
\newblock \emph{arXiv preprint arXiv:2207.05221}.

\bibitem[{Karpukhin et~al.(2020)Karpukhin, Oguz, Min, Lewis, Wu, Edunov, Chen,
  and Yih}]{karpukhin-etal-2020-dense}
Vladimir Karpukhin, Barlas Oguz, Sewon Min, Patrick Lewis, Ledell Wu, Sergey
  Edunov, Danqi Chen, and Wen-tau Yih. 2020.
\newblock \href {https://doi.org/10.18653/v1/2020.emnlp-main.550} {Dense
  passage retrieval for open-domain question answering}.
\newblock In \emph{Proceedings of the 2020 Conference on Empirical Methods in
  Natural Language Processing (EMNLP)}, pages 6769--6781, Online. Association
  for Computational Linguistics.

\bibitem[{K{\"o}ksal et~al.(2023)K{\"o}ksal, Schick, Korhonen, and
  Sch{\"u}tze}]{koksal2023longform}
Abdullatif K{\"o}ksal, Timo Schick, Anna Korhonen, and Hinrich Sch{\"u}tze.
  2023.
\newblock Longform: Optimizing instruction tuning for long text generation with
  corpus extraction.
\newblock \emph{arXiv preprint arXiv:2304.08460}.

\bibitem[{Kumari et~al.()Kumari, Shafqat, and Sarda}]{kumariretrieval}
Lilly Kumari, Usama~Bin Shafqat, and Nikhil Sarda.
\newblock Retrieval augmented generation for dialog modeling.

\bibitem[{Lee et~al.(2023)Lee, Hartmann, Park, Papailiopoulos, and
  Lee}]{lee-etal-2023-prompted}
Gibbeum Lee, Volker Hartmann, Jongho Park, Dimitris Papailiopoulos, and
  Kangwook Lee. 2023.
\newblock \href {https://doi.org/10.18653/v1/2023.findings-acl.277} {Prompted
  {LLM}s as chatbot modules for long open-domain conversation}.
\newblock In \emph{Findings of the Association for Computational Linguistics:
  ACL 2023}, pages 4536--4554, Toronto, Canada. Association for Computational
  Linguistics.

\bibitem[{Liang et~al.(2023)Liang, Tang, Li, and Zhang}]{liang-etal-2023-open}
Xiaobo Liang, Zecheng Tang, Juntao Li, and Min Zhang. 2023.
\newblock \href {https://doi.org/10.18653/v1/2023.acl-long.13} {Open-ended long
  text generation via masked language modeling}.
\newblock In \emph{Proceedings of the 61st Annual Meeting of the Association
  for Computational Linguistics (Volume 1: Long Papers)}, pages 223--241,
  Toronto, Canada. Association for Computational Linguistics.

\bibitem[{Lin(2004)}]{lin-2004-rouge}
Chin-Yew Lin. 2004.
\newblock \href {https://aclanthology.org/W04-1013} {{ROUGE}: A package for
  automatic evaluation of summaries}.
\newblock In \emph{Text Summarization Branches Out}, pages 74--81, Barcelona,
  Spain. Association for Computational Linguistics.

\bibitem[{Lin et~al.(2022)Lin, Hilton, and Evans}]{lin2022teaching}
Stephanie Lin, Jacob Hilton, and Owain Evans. 2022.
\newblock Teaching models to express their uncertainty in words.
\newblock \emph{arXiv preprint arXiv:2205.14334}.

\bibitem[{Liu et~al.(2023)Liu, Huang, and Chang}]{liu-etal-2023-ask}
Yuxiang Liu, Jie Huang, and Kevin Chang. 2023.
\newblock \href {https://doi.org/10.18653/v1/2023.findings-emnlp.178} {Ask to
  the point: Open-domain entity-centric question generation}.
\newblock In \emph{Findings of the Association for Computational Linguistics:
  EMNLP 2023}, pages 2703--2716, Singapore. Association for Computational
  Linguistics.

\bibitem[{Lu et~al.(2023)Lu, An, Lin, Pergola, He, Yin, Sun, and
  Wu}]{lu2023memochat}
Junru Lu, Siyu An, Mingbao Lin, Gabriele Pergola, Yulan He, Di~Yin, Xing Sun,
  and Yunsheng Wu. 2023.
\newblock Memochat: Tuning llms to use memos for consistent long-range
  open-domain conversation.
\newblock \emph{arXiv preprint arXiv:2308.08239}.

\bibitem[{Madaan et~al.(2024)Madaan, Tandon, Gupta, Hallinan, Gao, Wiegreffe,
  Alon, Dziri, Prabhumoye, Yang et~al.}]{madaan2024self}
Aman Madaan, Niket Tandon, Prakhar Gupta, Skyler Hallinan, Luyu Gao, Sarah
  Wiegreffe, Uri Alon, Nouha Dziri, Shrimai Prabhumoye, Yiming Yang, et~al.
  2024.
\newblock Self-refine: Iterative refinement with self-feedback.
\newblock \emph{Advances in Neural Information Processing Systems}, 36.

\bibitem[{Mao et~al.(2022)Mao, Wu, Ni, Zhang, Zhang, Yu, Deb, Zhu, Awadallah,
  and Radev}]{mao-etal-2022-dyle}
Ziming Mao, Chen~Henry Wu, Ansong Ni, Yusen Zhang, Rui Zhang, Tao Yu,
  Budhaditya Deb, Chenguang Zhu, Ahmed Awadallah, and Dragomir Radev. 2022.
\newblock \href {https://doi.org/10.18653/v1/2022.acl-long.118} {{DYLE}:
  Dynamic latent extraction for abstractive long-input summarization}.
\newblock In \emph{Proceedings of the 60th Annual Meeting of the Association
  for Computational Linguistics (Volume 1: Long Papers)}, pages 1687--1698,
  Dublin, Ireland. Association for Computational Linguistics.

\bibitem[{Min et~al.(2023)Min, Krishna, Lyu, Lewis, Yih, Koh, Iyyer,
  Zettlemoyer, and Hajishirzi}]{min-etal-2023-factscore}
Sewon Min, Kalpesh Krishna, Xinxi Lyu, Mike Lewis, Wen-tau Yih, Pang Koh, Mohit
  Iyyer, Luke Zettlemoyer, and Hannaneh Hajishirzi. 2023.
\newblock \href {https://doi.org/10.18653/v1/2023.emnlp-main.741}
  {{FA}ct{S}core: Fine-grained atomic evaluation of factual precision in long
  form text generation}.
\newblock In \emph{Proceedings of the 2023 Conference on Empirical Methods in
  Natural Language Processing}, pages 12076--12100, Singapore. Association for
  Computational Linguistics.

\bibitem[{Mirowski et~al.(2023)Mirowski, Mathewson, Pittman, and
  Evans}]{mirowski2023co}
Piotr Mirowski, Kory~W Mathewson, Jaylen Pittman, and Richard Evans. 2023.
\newblock Co-writing screenplays and theatre scripts with language models:
  Evaluation by industry professionals.
\newblock In \emph{Proceedings of the 2023 CHI Conference on Human Factors in
  Computing Systems}, pages 1--34.

\bibitem[{Moryossef et~al.(2019)Moryossef, Goldberg, and
  Dagan}]{moryossef-etal-2019-step}
Amit Moryossef, Yoav Goldberg, and Ido Dagan. 2019.
\newblock \href {https://doi.org/10.18653/v1/N19-1236} {{S}tep-by-step:
  {S}eparating planning from realization in neural data-to-text generation}.
\newblock In \emph{Proceedings of the 2019 Conference of the North {A}merican
  Chapter of the Association for Computational Linguistics: Human Language
  Technologies, Volume 1 (Long and Short Papers)}, pages 2267--2277,
  Minneapolis, Minnesota. Association for Computational Linguistics.

\bibitem[{Narayan et~al.(2023)Narayan, Maynez, Amplayo, Ganchev, Louis, Huot,
  Sandholm, Das, and Lapata}]{narayan-etal-2023-conditional}
Shashi Narayan, Joshua Maynez, Reinald~Kim Amplayo, Kuzman Ganchev, Annie
  Louis, Fantine Huot, Anders Sandholm, Dipanjan Das, and Mirella Lapata. 2023.
\newblock \href {https://doi.org/10.1162/tacl_a_00583} {Conditional generation
  with a question-answering blueprint}.
\newblock \emph{Transactions of the Association for Computational Linguistics},
  11:974--996.

\bibitem[{Papineni et~al.(2002)Papineni, Roukos, Ward, and
  Zhu}]{papineni-etal-2002-bleu}
Kishore Papineni, Salim Roukos, Todd Ward, and Wei-Jing Zhu. 2002.
\newblock \href {https://doi.org/10.3115/1073083.1073135} {{B}leu: a method for
  automatic evaluation of machine translation}.
\newblock In \emph{Proceedings of the 40th Annual Meeting of the Association
  for Computational Linguistics}, pages 311--318, Philadelphia, Pennsylvania,
  USA. Association for Computational Linguistics.

\bibitem[{Ram et~al.(2023{\natexlab{a}})Ram, Levine, Dalmedigos, Muhlgay,
  Shashua, Leyton-Brown, and Shoham}]{ram-etal-2023-context}
Ori Ram, Yoav Levine, Itay Dalmedigos, Dor Muhlgay, Amnon Shashua, Kevin
  Leyton-Brown, and Yoav Shoham. 2023{\natexlab{a}}.
\newblock \href {https://doi.org/10.1162/tacl_a_00605} {In-context
  retrieval-augmented language models}.
\newblock \emph{Transactions of the Association for Computational Linguistics},
  11:1316--1331.

\bibitem[{Ram et~al.(2023{\natexlab{b}})Ram, Levine, Dalmedigos, Muhlgay,
  Shashua, Leyton-Brown, and Shoham}]{ram2023context}
Ori Ram, Yoav Levine, Itay Dalmedigos, Dor Muhlgay, Amnon Shashua, Kevin
  Leyton-Brown, and Yoav Shoham. 2023{\natexlab{b}}.
\newblock In-context retrieval-augmented language models.
\newblock \emph{Transactions of the Association for Computational Linguistics},
  11:1316--1331.

\bibitem[{Rawte et~al.(2023)Rawte, Sheth, and Das}]{rawte2023survey}
Vipula Rawte, Amit Sheth, and Amitava Das. 2023.
\newblock A survey of hallucination in large foundation models.
\newblock \emph{arXiv preprint arXiv:2309.05922}.

\bibitem[{Shao et~al.(2024{\natexlab{a}})Shao, Jiang, Kanell, Xu, Khattab, and
  Lam}]{shao-etal-2024-assisting}
Yijia Shao, Yucheng Jiang, Theodore Kanell, Peter Xu, Omar Khattab, and Monica
  Lam. 2024{\natexlab{a}}.
\newblock \href {https://doi.org/10.18653/v1/2024.naacl-long.347} {Assisting in
  writing {W}ikipedia-like articles from scratch with large language models}.
\newblock In \emph{Proceedings of the 2024 Conference of the North American
  Chapter of the Association for Computational Linguistics: Human Language
  Technologies (Volume 1: Long Papers)}, pages 6252--6278, Mexico City, Mexico.
  Association for Computational Linguistics.

\bibitem[{Shao et~al.(2024{\natexlab{b}})Shao, Jiang, Kanell, Xu, Khattab, and
  Lam}]{shao2024assisting}
Yijia Shao, Yucheng Jiang, Theodore~A Kanell, Peter Xu, Omar Khattab, and
  Monica~S Lam. 2024{\natexlab{b}}.
\newblock Assisting in writing wikipedia-like articles from scratch with large
  language models.
\newblock \emph{arXiv preprint arXiv:2402.14207}.

\bibitem[{Shen et~al.(2019)Shen, Celikyilmaz, Zhang, Chen, Wang, Gao, and
  Carin}]{shen-etal-2019-towards}
Dinghan Shen, Asli Celikyilmaz, Yizhe Zhang, Liqun Chen, Xin Wang, Jianfeng
  Gao, and Lawrence Carin. 2019.
\newblock \href {https://doi.org/10.18653/v1/P19-1200} {Towards generating long
  and coherent text with multi-level latent variable models}.
\newblock In \emph{Proceedings of the 57th Annual Meeting of the Association
  for Computational Linguistics}, pages 2079--2089, Florence, Italy.
  Association for Computational Linguistics.

\bibitem[{Shi et~al.(2023)Shi, Min, Yasunaga, Seo, James, Lewis, Zettlemoyer,
  and Yih}]{shi2023replug}
Weijia Shi, Sewon Min, Michihiro Yasunaga, Minjoon Seo, Rich James, Mike Lewis,
  Luke Zettlemoyer, and Wen-tau Yih. 2023.
\newblock Replug: Retrieval-augmented black-box language models.
\newblock \emph{arXiv preprint arXiv:2301.12652}.

\bibitem[{Slobodkin et~al.(2023)Slobodkin, Goldman, Caciularu, Dagan, and
  Ravfogel}]{slobodkin-etal-2023-curious}
Aviv Slobodkin, Omer Goldman, Avi Caciularu, Ido Dagan, and Shauli Ravfogel.
  2023.
\newblock \href {https://doi.org/10.18653/v1/2023.emnlp-main.220} {The curious
  case of hallucinatory (un)answerability: Finding truths in the hidden states
  of over-confident large language models}.
\newblock In \emph{Proceedings of the 2023 Conference on Empirical Methods in
  Natural Language Processing}, pages 3607--3625, Singapore. Association for
  Computational Linguistics.

\bibitem[{Sun et~al.(2022)Sun, Sun, Meng, Li, and
  Fan}]{sun-etal-2022-summarize}
Xiaofei Sun, Zijun Sun, Yuxian Meng, Jiwei Li, and Chun Fan. 2022.
\newblock \href {https://aclanthology.org/2022.coling-1.556} {Summarize,
  outline, and elaborate: Long-text generation via hierarchical supervision
  from extractive summaries}.
\newblock In \emph{Proceedings of the 29th International Conference on
  Computational Linguistics}, pages 6392--6402, Gyeongju, Republic of Korea.
  International Committee on Computational Linguistics.

\bibitem[{Tian et~al.(2023)Tian, Mitchell, Zhou, Sharma, Rafailov, Yao, Finn,
  and Manning}]{tian-etal-2023-just}
Katherine Tian, Eric Mitchell, Allan Zhou, Archit Sharma, Rafael Rafailov,
  Huaxiu Yao, Chelsea Finn, and Christopher Manning. 2023.
\newblock \href {https://doi.org/10.18653/v1/2023.emnlp-main.330} {Just ask for
  calibration: Strategies for eliciting calibrated confidence scores from
  language models fine-tuned with human feedback}.
\newblock In \emph{Proceedings of the 2023 Conference on Empirical Methods in
  Natural Language Processing}, pages 5433--5442, Singapore. Association for
  Computational Linguistics.

\bibitem[{Vijayakumar(2024)}]{vijayakumar2024exemplification}
Chintalapalli Vijayakumar. 2024.
\newblock Exemplification in student essay writing: A study of learner corpus
  of essay writing (lcew).
\newblock \emph{International Journal of Applied Linguistics},
  34(4):1514--1532.

\bibitem[{Wang et~al.(2023)Wang, Ding, Cao, Tian, Wang, Tao, and
  Guo}]{wang2023recursively}
Qingyue Wang, Liang Ding, Yanan Cao, Zhiliang Tian, Shi Wang, Dacheng Tao, and
  Li~Guo. 2023.
\newblock Recursively summarizing enables long-term dialogue memory in large
  language models.
\newblock \emph{arXiv preprint arXiv:2308.15022}.

\bibitem[{Wette(2014)}]{wette2014teachers}
Rosemary Wette. 2014.
\newblock Teachers' practices in eap writing instruction: Use of models and
  modeling.
\newblock \emph{System}, 42:60--69.

\bibitem[{Wu(2019)}]{wu2019understanding}
Zhiwei Wu. 2019.
\newblock Understanding students’ mimicry, emulation and imitation of genre
  exemplars: An exploratory study.
\newblock \emph{English for Specific Purposes}, 54:127--138.

\bibitem[{Xiong et~al.(2023)Xiong, Hu, Lu, Li, Fu, He, and Hooi}]{xiong2023can}
Miao Xiong, Zhiyuan Hu, Xinyang Lu, Yifei Li, Jie Fu, Junxian He, and Bryan
  Hooi. 2023.
\newblock Can llms express their uncertainty? an empirical evaluation of
  confidence elicitation in llms.
\newblock \emph{arXiv preprint arXiv:2306.13063}.

\bibitem[{Xu et~al.(2024)Xu, Wang, and Chen}]{xu2024towards}
Han Xu, Xingyuan Wang, and Haipeng Chen. 2024.
\newblock Towards real-time and personalized code generation.
\newblock In \emph{Proceedings of the 33rd ACM International Conference on
  Information and Knowledge Management}, pages 5568--5569.

\bibitem[{Yamada et~al.(2020)Yamada, Asai, Sakuma, Shindo, Takeda, Takefuji,
  and Matsumoto}]{yamada-etal-2020-wikipedia2vec}
Ikuya Yamada, Akari Asai, Jin Sakuma, Hiroyuki Shindo, Hideaki Takeda,
  Yoshiyasu Takefuji, and Yuji Matsumoto. 2020.
\newblock \href {https://doi.org/10.18653/v1/2020.emnlp-demos.4}
  {{W}ikipedia2{V}ec: An efficient toolkit for learning and visualizing the
  embeddings of words and entities from {W}ikipedia}.
\newblock In \emph{Proceedings of the 2020 Conference on Empirical Methods in
  Natural Language Processing: System Demonstrations}, pages 23--30, Online.
  Association for Computational Linguistics.

\bibitem[{Yang et~al.(2023{\natexlab{a}})Yang, Klein, Peng, and
  Tian}]{yang-etal-2023-doc}
Kevin Yang, Dan Klein, Nanyun Peng, and Yuandong Tian. 2023{\natexlab{a}}.
\newblock \href {https://doi.org/10.18653/v1/2023.acl-long.190} {{DOC}:
  Improving long story coherence with detailed outline control}.
\newblock In \emph{Proceedings of the 61st Annual Meeting of the Association
  for Computational Linguistics (Volume 1: Long Papers)}, pages 3378--3465,
  Toronto, Canada. Association for Computational Linguistics.

\bibitem[{Yang et~al.(2022)Yang, Tian, Peng, and Klein}]{yang-etal-2022-re3}
Kevin Yang, Yuandong Tian, Nanyun Peng, and Dan Klein. 2022.
\newblock \href {https://doi.org/10.18653/v1/2022.emnlp-main.296} {Re3:
  Generating longer stories with recursive reprompting and revision}.
\newblock In \emph{Proceedings of the 2022 Conference on Empirical Methods in
  Natural Language Processing}, pages 4393--4479, Abu Dhabi, United Arab
  Emirates. Association for Computational Linguistics.

\bibitem[{Yang et~al.(2023{\natexlab{b}})Yang, Chern, Qiu, Neubig, and
  Liu}]{yang2023alignment}
Yuqing Yang, Ethan Chern, Xipeng Qiu, Graham Neubig, and Pengfei Liu.
  2023{\natexlab{b}}.
\newblock Alignment for honesty.
\newblock \emph{arXiv preprint arXiv:2312.07000}.

\bibitem[{You et~al.(2023)You, Wu, Liang, Mao, Wu, Cao, Cai, Guo, Xia, Wei
  et~al.}]{you2023eipe}
Wang You, Wenshan Wu, Yaobo Liang, Shaoguang Mao, Chenfei Wu, Maosong Cao,
  Yuzhe Cai, Yiduo Guo, Yan Xia, Furu Wei, et~al. 2023.
\newblock Eipe-text: Evaluation-guided iterative plan extraction for long-form
  narrative text generation.
\newblock \emph{arXiv preprint arXiv:2310.08185}.

\bibitem[{Zhang et~al.(2023)Zhang, Diao, Lin, Fung, Lian, Wang, Chen, Ji, and
  Zhang}]{zhang2023r}
Hanning Zhang, Shizhe Diao, Yong Lin, Yi~R Fung, Qing Lian, Xingyao Wang,
  Yangyi Chen, Heng Ji, and Tong Zhang. 2023.
\newblock R-tuning: Teaching large language models to refuse unknown questions.
\newblock \emph{arXiv preprint arXiv:2311.09677}.

\bibitem[{Zhang et~al.(2019)Zhang, Guo, Fan, Lan, and Cheng}]{zhang2019outline}
Ruqing Zhang, Jiafeng Guo, Yixing Fan, Yanyan Lan, and Xueqi Cheng. 2019.
\newblock Outline generation: Understanding the inherent content structure of
  documents.
\newblock In \emph{Proceedings of the 42nd International ACM SIGIR Conference
  on Research and Development in Information Retrieval}, pages 745--754.

\bibitem[{Zhang et~al.(2022)Zhang, Ni, Mao, Wu, Zhu, Deb, Awadallah, Radev, and
  Zhang}]{zhang-etal-2022-summn}
Yusen Zhang, Ansong Ni, Ziming Mao, Chen~Henry Wu, Chenguang Zhu, Budhaditya
  Deb, Ahmed Awadallah, Dragomir Radev, and Rui Zhang. 2022.
\newblock \href {https://doi.org/10.18653/v1/2022.acl-long.112} {{S}umm$^n$: A
  multi-stage summarization framework for long input dialogues and documents}.
\newblock In \emph{Proceedings of the 60th Annual Meeting of the Association
  for Computational Linguistics (Volume 1: Long Papers)}, pages 1592--1604,
  Dublin, Ireland. Association for Computational Linguistics.

\bibitem[{Zheng et~al.(2024)Zheng, Chiang, Sheng, Zhuang, Wu, Zhuang, Lin, Li,
  Li, Xing et~al.}]{zheng2024judging}
Lianmin Zheng, Wei-Lin Chiang, Ying Sheng, Siyuan Zhuang, Zhanghao Wu, Yonghao
  Zhuang, Zi~Lin, Zhuohan Li, Dacheng Li, Eric Xing, et~al. 2024.
\newblock Judging llm-as-a-judge with mt-bench and chatbot arena.
\newblock \emph{Advances in Neural Information Processing Systems}, 36.

\bibitem[{Zhou et~al.(2023)Zhou, Jiang, Cui, Wang, Xiao, Hou, Cotterell, and
  Sachan}]{zhou2023recurrentgpt}
Wangchunshu Zhou, Yuchen~Eleanor Jiang, Peng Cui, Tiannan Wang, Zhenxin Xiao,
  Yifan Hou, Ryan Cotterell, and Mrinmaya Sachan. 2023.
\newblock Recurrentgpt: Interactive generation of (arbitrarily) long text.
\newblock \emph{arXiv preprint arXiv:2305.13304}.

\end{thebibliography}

\appendix
\section{A Complete Example}\label{appendix:example}

\begin{figure*}[tp!]
\centerline{\includegraphics[width=1\linewidth]{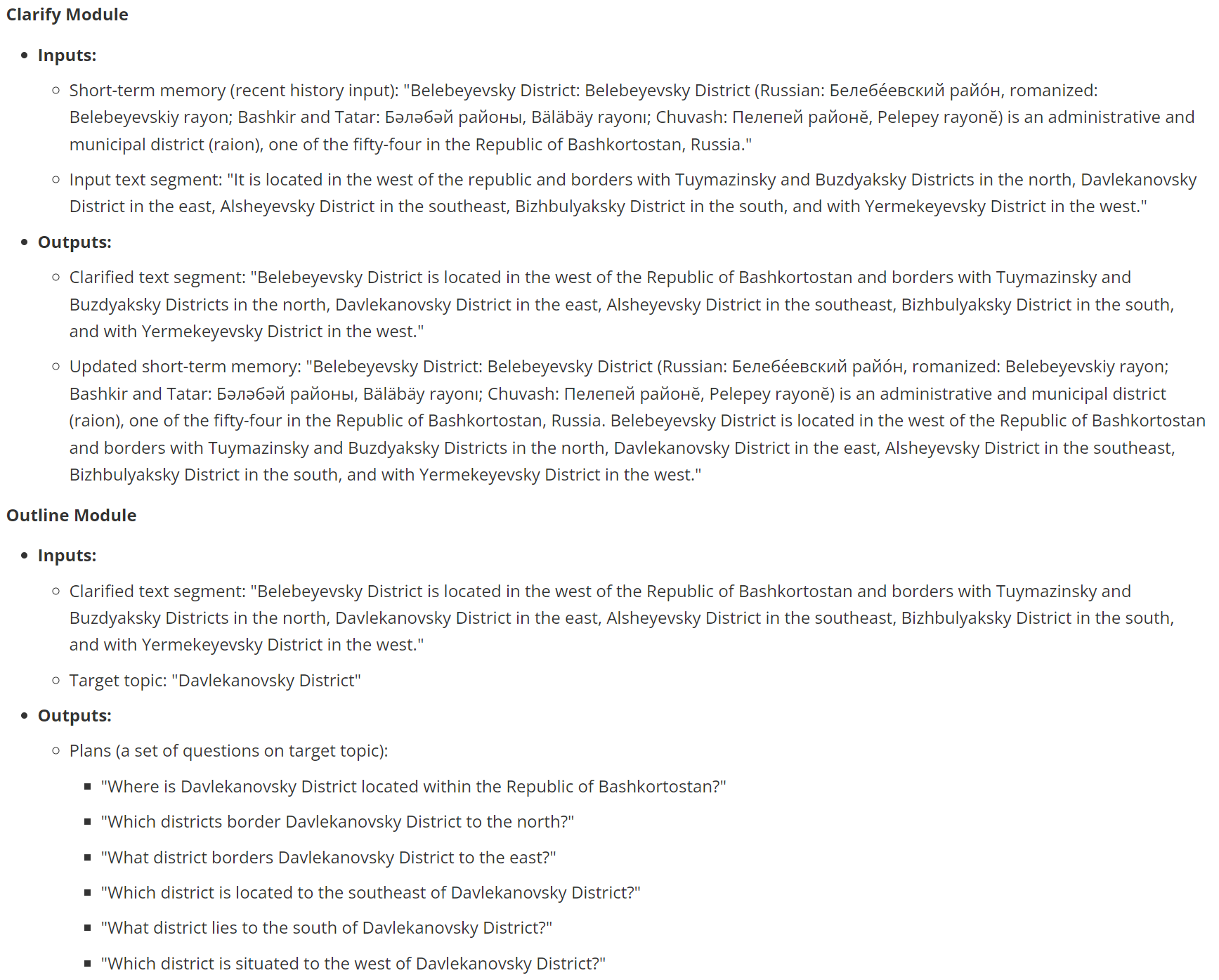}}
\centerline{\includegraphics[width=1\linewidth]{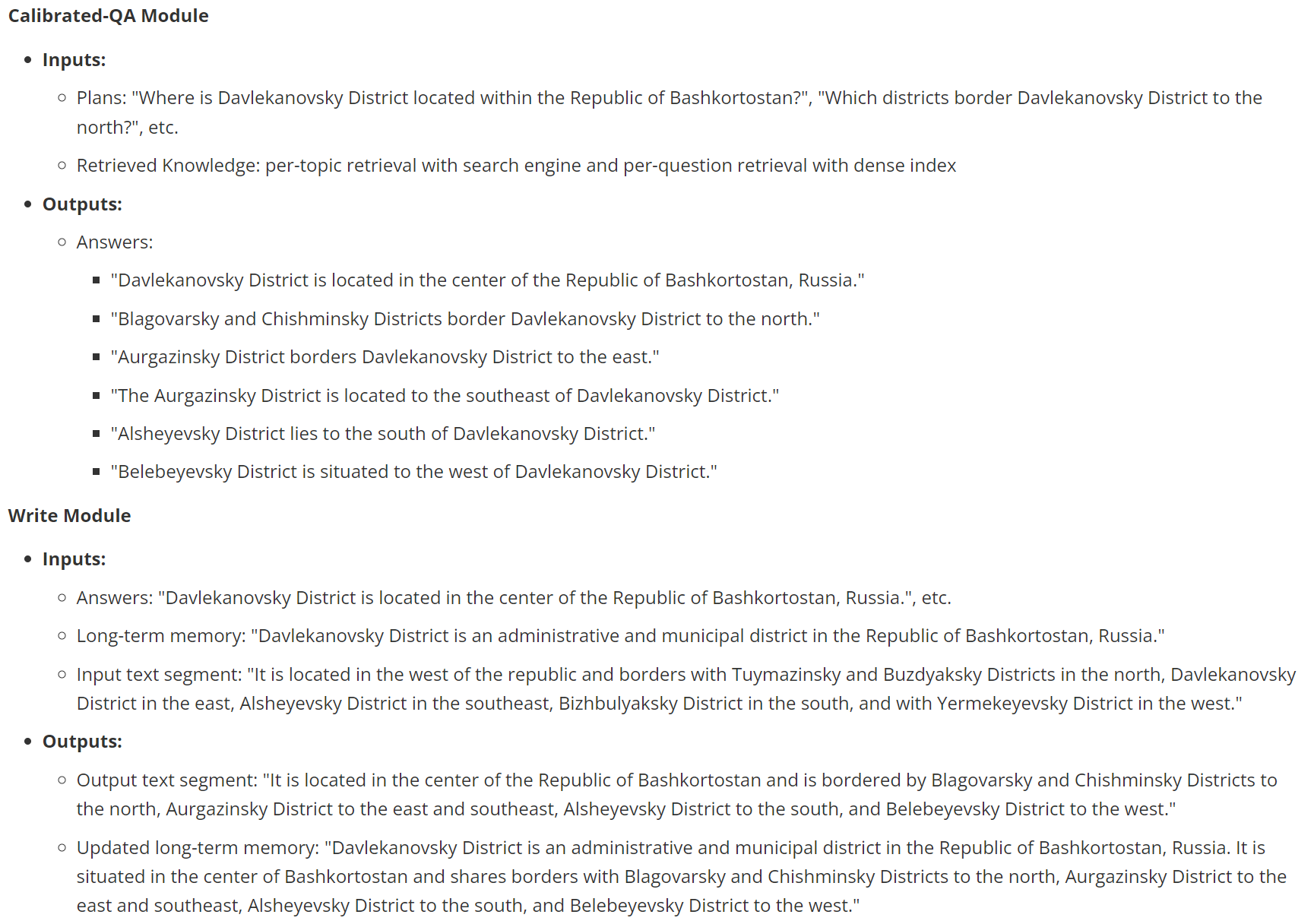}}
\caption{A complete running example.}
\label{fig:runningexample}
\end{figure*}

We present a complete running example of the inputs and outputs for each module in a recurrent step in Figure~\ref{fig:runningexample}.

\section{Prompts for Model}\label{appendix:prompts-model}

\begin{figure*}[tp!]
\centerline{\includegraphics[width=0.85\linewidth]{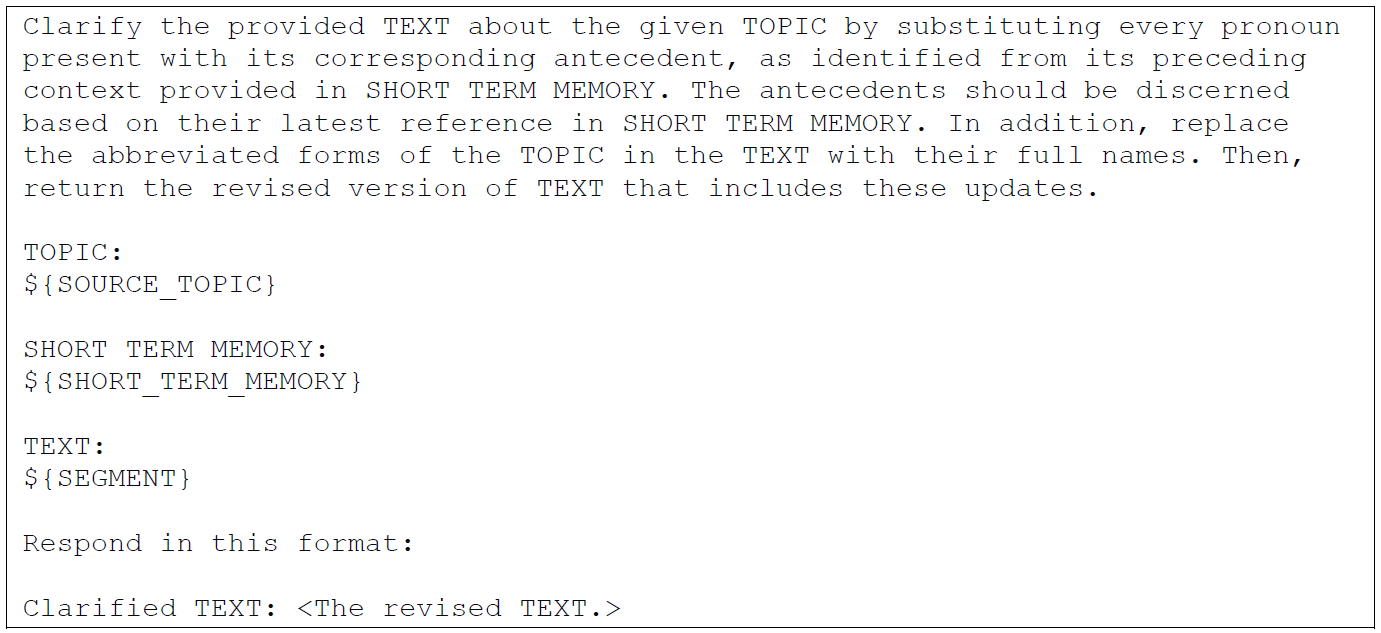}}
\caption{Prompt of the Clarify component in \textsc{Plan} stage.}
\label{fig:prompt-clarify}
\end{figure*}

\begin{figure*}[tp!]
\centerline{\includegraphics[width=0.85\linewidth]{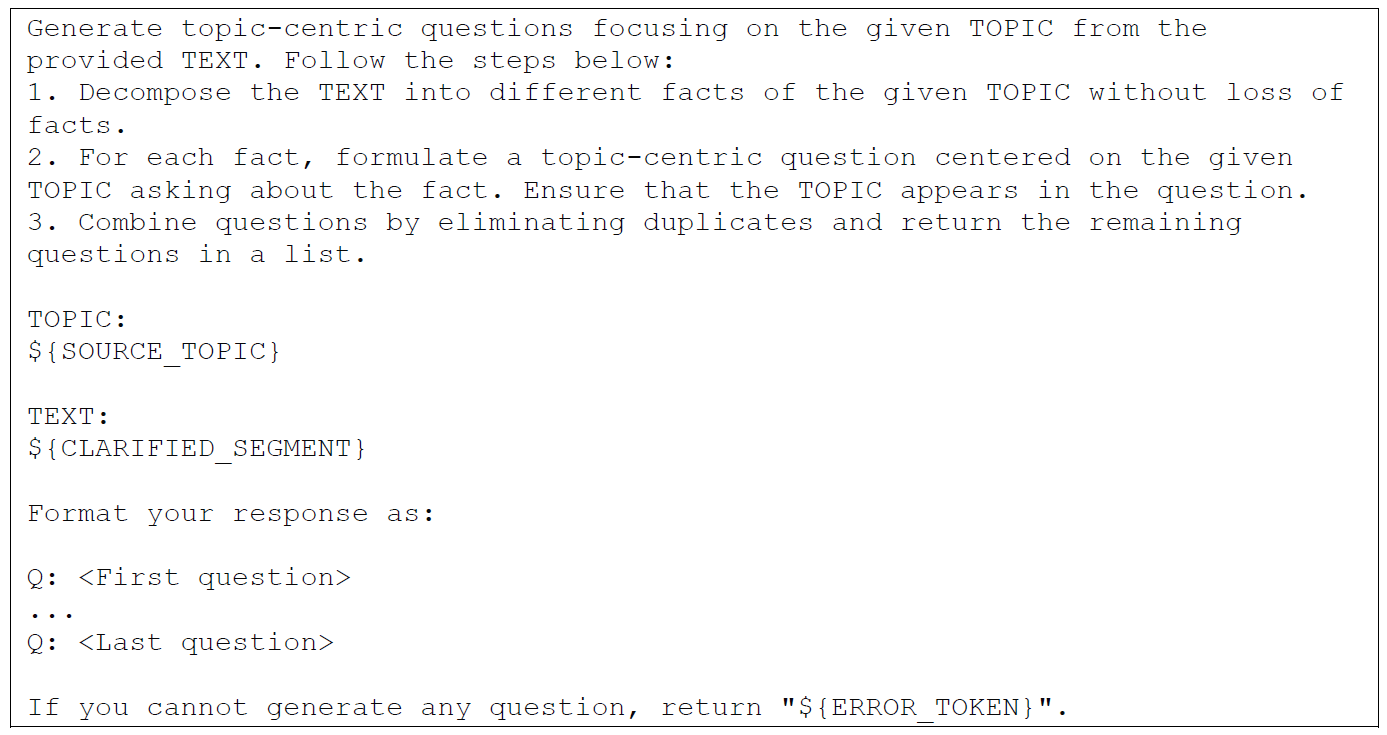}}
\caption{Prompt of the Outline component in \textsc{Plan} stage.}
\label{fig:prompt-outline}
\end{figure*}

\begin{figure*}[tp!]
\centerline{\includegraphics[width=0.85\linewidth]{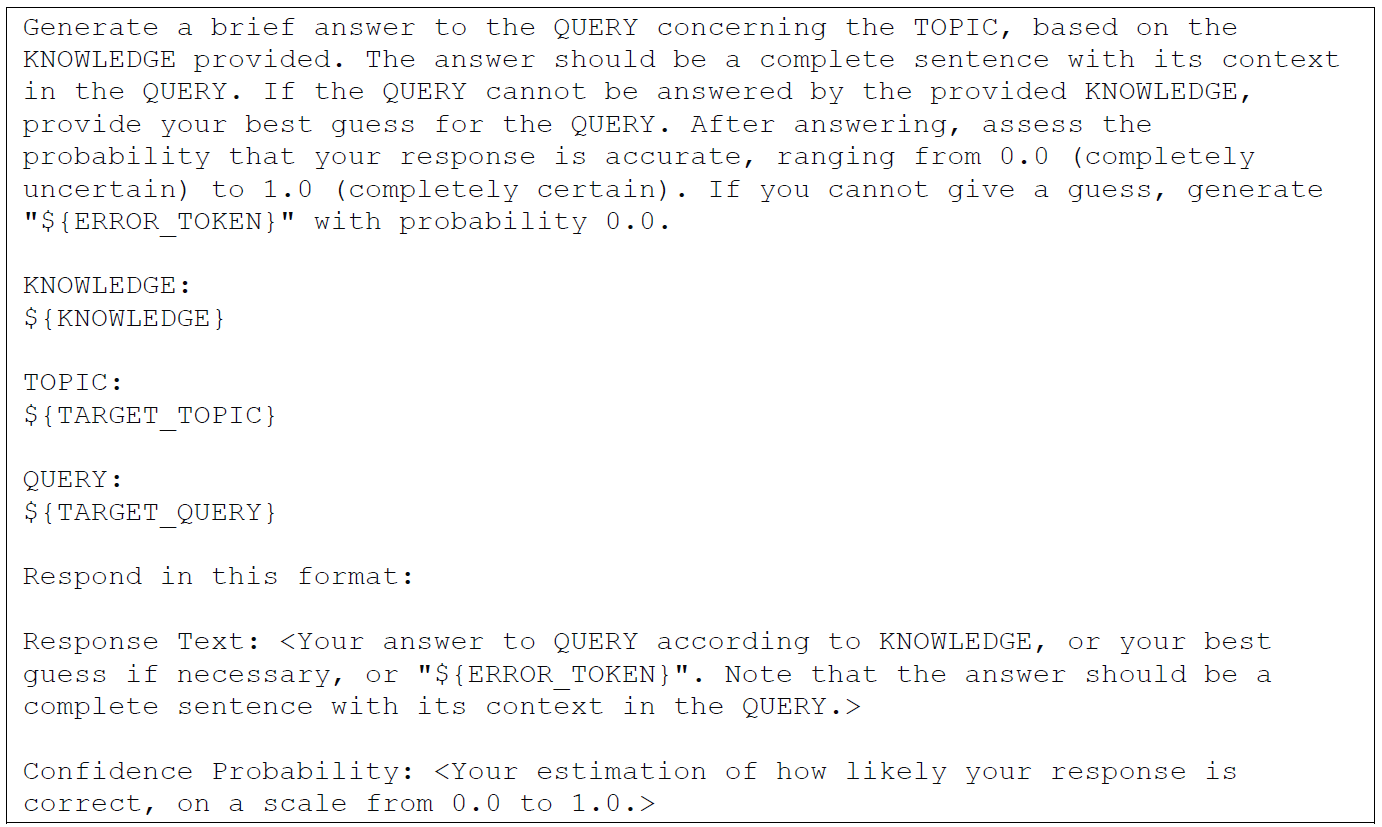}}
\caption{Prompt of the Calibrated-QA component in \textsc{Adapt} stage.}
\label{fig:prompt-calibratedQA}
\end{figure*}

\begin{figure*}[tp!]
\centerline{\includegraphics[width=0.9\linewidth]{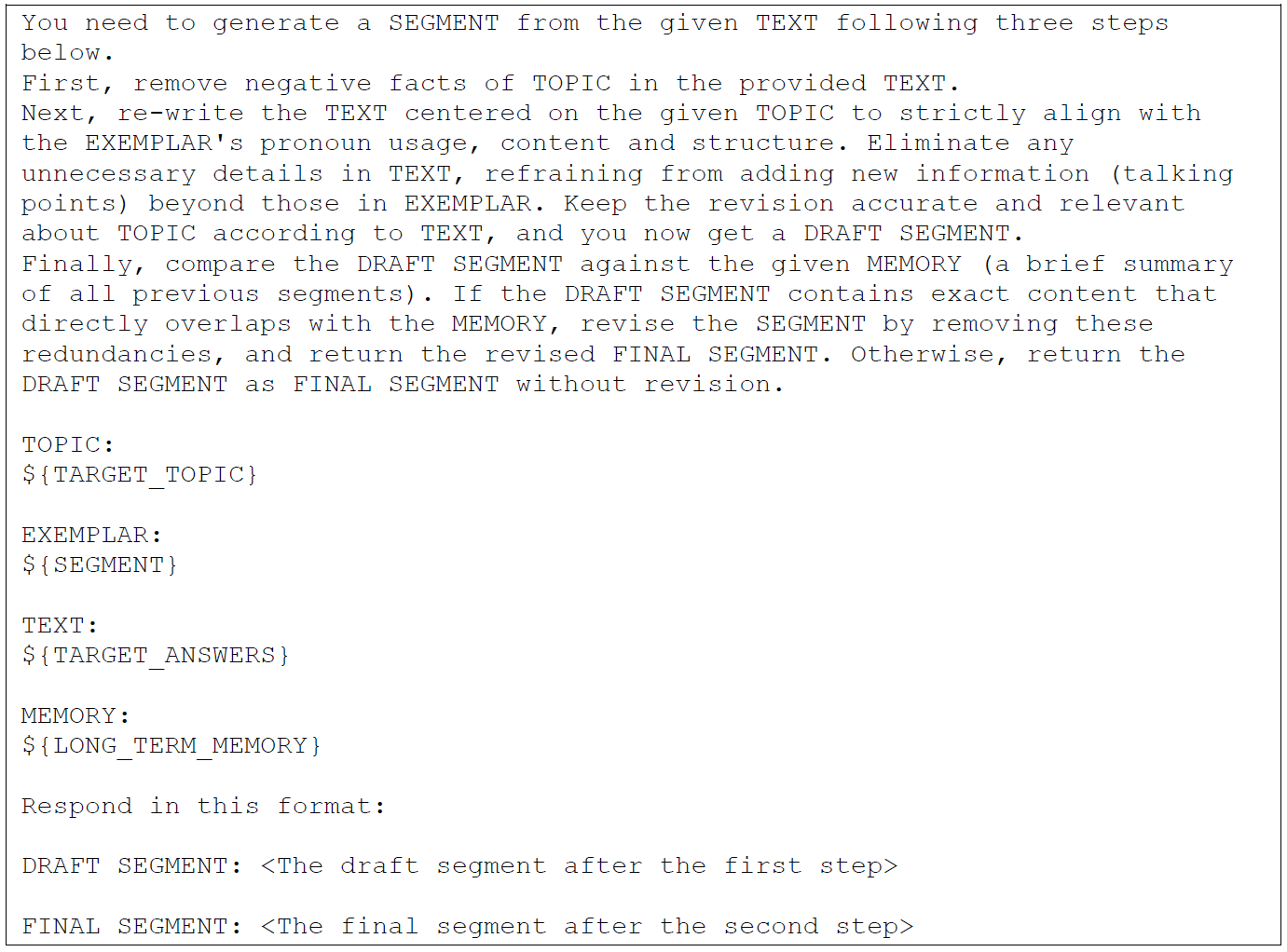}}
\caption{Prompt of the Write component in \textsc{Adapt} stage.}
\label{fig:prompt-write}
\end{figure*}

\begin{figure*}[tp!]
\centerline{\includegraphics[width=0.9\linewidth]{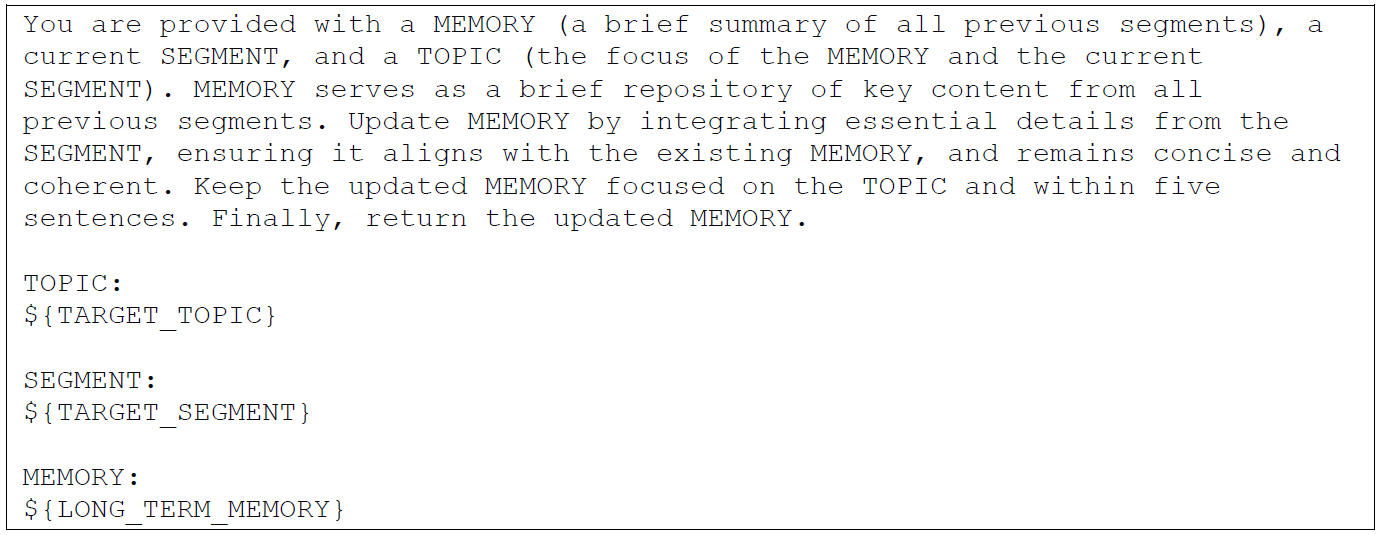}}
\caption{Prompt of the post-Write summarization step in \textsc{Adapt} stage.}
\label{fig:prompt-summarize}
\end{figure*}

In our proposed \textsc{RePA} model, we use five prompts in each \textsc{Plan-then-Adapt} recurrent step to achieve fine-grained control of LLMs: the Clarify prompt (Figure~\ref{fig:prompt-clarify}), Outline prompt (Figure~\ref{fig:prompt-outline}), Calibrated-QA prompt (Figure~\ref{fig:prompt-calibratedQA}), Write prompt (Figure~\ref{fig:prompt-write}), and Summarize prompt (Figure~\ref{fig:prompt-summarize}). 

\section{Case Study on Longer Texts}\label{appendix:longer-texts}

Given the lack of extensive, high-quality, paired longer-text datasets, we evaluated our model's capacity for generating extended outputs using a manually curated example. Specifically, we tasked the model with generating a lengthy text on a target topic, "Beyoncé" (Table~\ref{tab:longtext-output}) based on an exemplar on "Taylor Swift" (Table~\ref{tab:longtext-source}). In future research, larger-scale evaluations with additional longer-text datasets could provide more comprehensive assessments.

\begin{table*}[htbp]
\small
\centering

\doublespacing
\spaceskip=3pt
\setlength{\parskip}{15pt}

\begin{tabular}{|p{15.5cm}|}
\hline

\quad\quad Taylor Alison Swift (born December 13, 1989) is an American singer-songwriter and director. Taylor Swift is regarded as an influential cultural figure of the 21st century. Taylor Swift is regarded as an influential cultural figure of the 21st century. Throughout her career, Taylor Swift has been recognized for her heartfelt lyrics and catchy melodies. She rose to fame following the release of her self-titled debut album in October 2006. \\

\quad\quad Starting her career as a solo artist, Taylor Alison Swift has achieved global superstardom, including winning the Grammy Award for album of the year for Midnights (2022), suggesting she is among the best-selling artists of all time. Taylor Swift's self-titled debut album was released in October 2006. She then followed with the U.S. number-one solo albums "Taylor Swift" (2006), "Fearless" (2008), and "Speak Now" (2011). After creating her own management company, 13 Management, Taylor Swift achieved critical acclaim for her self-titled debut album "Taylor Swift," which explored themes such as love, dreams, and personal experiences as a teenager. 

\quad\quad Taylor Swift's most successful songs on the Billboard Hot 100 include her numerous hits that have defined her career.  Outside of music, Taylor Swift has starred as an actress in films such as The Pink Panther (2006), Obsessed (2009), and The Lion King (2019). Taylor Swift has made a significant name for herself in the music industry. Her accolades include a record 32 Grammy Awards (including the 2010 Song of the Year), as well as 26 MTV Video Music Awards (including the 2014 Michael Jackson Video Vanguard Award) – all of which are more than any other artist in the music industry. She is known for her significant impact on contemporary music and her successful tours. \\

\quad\quad Taylor Alison Swift was born on December 13, 1989, at the hospital in West Reading, Pennsylvania to Andrea Gardner Swift (née Finlay), a mutual fund marketing executive, and Scott Kingsley Swift, a stockbroker for Merrill Lynch. Taylor Swift's mother, Andrea Gardner Swift, worked in mutual fund marketing, and her father, Scott Kingsley Swift, was a stockbroker. Taylor Swift is a distinguished artist with no familial ties to Giselle Knowles-Carter or Solange Knowles. Taylor Swift's immediate family includes her parents, Scott Kingsley Swift and Andrea Gardner Swift. \\

\quad\quad Taylor Swift's interest in music and performing was evident from a young age, showcasing her talents in a school talent show. Taylor Swift's voice captivates audiences with her heartfelt lyrics and catchy melodies, marking her as a significant figure in contemporary music. Her vocal abilities distinguish her as a significant figure in contemporary music. \\

\quad\quad Taylor Swift's music is generally pop and country, captivating audiences with her heartfelt lyrics and catchy melodies. Taylor Swift almost exclusively releases her songs in English. Taylor Swift's early career was centered on country music, showcasing her storytelling abilities and personal experiences through her songs. She then transitioned to pop music, marking a significant shift in her musical style and broadening her appeal to a global audience. She is tied with American lyricist Diane Warren at third with nine songwriting credits on number-one singles. The controversy surrounding Taylor Swift's songwriting credits began with interviews in which she attributed herself as the songwriter for songs in which she was a co-writer or for which her contributions were marginal. \\

\quad\quad Taylor Swift's global superstardom and history of winning prestigious awards, such as the Grammy for album of the year for Midnights in 2024, showcase her strong and dedicated fan base that likely contributes to her high sales figures. Taylor Swift's success has made her a cultural icon, captivating audiences with heartfelt lyrics and catchy melodies. She has become one of the most influential artists in contemporary music. She was the highest-paid musician of 2016, underscoring her significant influence in the music industry. \\

\quad\quad Writing for a leading entertainment outlet, it could be said that Taylor Swift has been a defining figure in music since 2010. Her approach to narrative songwriting and personal connection with audiences showcases the power of storytelling in music. Taylor Swift has been credited with reviving the album as an art form in an era dominated by singles and streaming. \\

\hline
\end{tabular}
\caption{\textsc{RePA}'s source text on "Taylor Swift".}
\label{tab:longtext-source}
\end{table*}

\begin{table*}[htbp]
\small
\centering

\doublespacing
\spaceskip=3pt
\setlength{\parskip}{15pt}

\begin{tabular}{|p{15.5cm}|}
\hline

\quad\quad Beyoncé Giselle Knowles-Carter (born September 4, 1981) is an American singer-songwriter and businesswoman. Beyoncé is regarded as an influential cultural figure of the 21st century. Throughout her career, Beyoncé has been recognized for her distinctive vocal range and live performances. She rose to fame in the late 1990s as the lead singer of Destiny's Child. \\

\quad\quad Starting her career in the music industry, Beyoncé formed the singing-rapping girl group Destiny’s Child in 1990 with childhood friends. She has achieved global superstardom, recognized for her distinctive vocal range, live performances, and as an influential cultural figure of the 21st century. Her success extends to music, business, and a significant social media presence with millions of followers. It is reasonable to infer that she is among the best-selling artists of all time. She then followed with the U.S. number-one solo albums "B'Day" (2006) and "4" (2011). After achieving fame as the lead singer of Destiny's Child, Beyoncé embarked on a solo career with her self-titled debut album "Beyoncé." \\

\quad\quad Beyoncé's most successful songs on the Billboard Hot 100 include her numerous hits that have defined her career. Beyoncé has made a significant name for herself in the music industry. Her accolades include the 2010 Song of the Year Grammy Award for "Single Ladies (Put a Ring on It)." She is known for her significant impact on contemporary music and her successful tours. \\

\quad\quad Beyoncé Giselle Knowles-Carter was born on September 4, 1981, at the hospital in Houston, Texas to Tina Knowles, a businesswoman in the fashion industry, and Mathew Knowles, involved in her music career. Beyoncé's mother, Tina Knowles, worked in the fashion industry, and her father, Mathew Knowles, was involved in her music career. Beyoncé's immediate family includes her parents, Tina Knowles (née Beyincé) and Mathew Knowles. \\

\quad\quad Beyoncé's interest in music and performing was evident from a young age, showcasing her talents by forming the singing-rapping girl group Destiny’s Child with childhood friends when she was nine years old. Beyoncé's voice captivates audiences with her distinctive vocal range and live performances, marking her as a significant figure in contemporary music. Her vocal abilities distinguish her as a significant figure in contemporary music. \\

\quad\quad Beyoncé's music is generally pop, R\&B, and hip hop, captivating audiences with her heartfelt lyrics and catchy melodies. Beyoncé almost exclusively releases her songs in English. Beyoncé's early career focused on R\&B and pop music, highlighting her storytelling skills and personal experiences in her songs. She transitioned to pop music, broadening her appeal to a global audience. She is tied with American lyricist Diane Warren at third with nine songwriting credits on number-one singles. \\

\quad\quad Beyoncé's history of winning prestigious awards, such as the Grammy Awards, showcases her global superstardom and strong, dedicated fan base that likely contributes to her high sales figures. Beyoncé's success has made her a cultural icon, captivating audiences with her distinctive vocal range and live performances. She has become one of the most influential artists in contemporary music. She was the highest-paid musician of 2016, underscoring her significant influence in the music industry. \\

\quad\quad Writing for a leading entertainment outlet, it could be said that Beyoncé has been a defining figure in music since 2010. Her approach to narrative songwriting and personal connection with audiences showcases the power of storytelling in music. Beyoncé has effectively navigated the shift towards streaming and singles, maintaining her relevance and success in the music industry. \\

\hline
\end{tabular}
\caption{\textsc{RePA}'s generated text for "Beyoncé".}
\label{tab:longtext-output}
\end{table*}
\section{Details on Task-Specific LLM-Judge}\label{appendix:LLMmetrics}

\subsection{Implementation Details}

\begin{figure*}[t!]
\centerline{\includegraphics[width=0.85\linewidth]{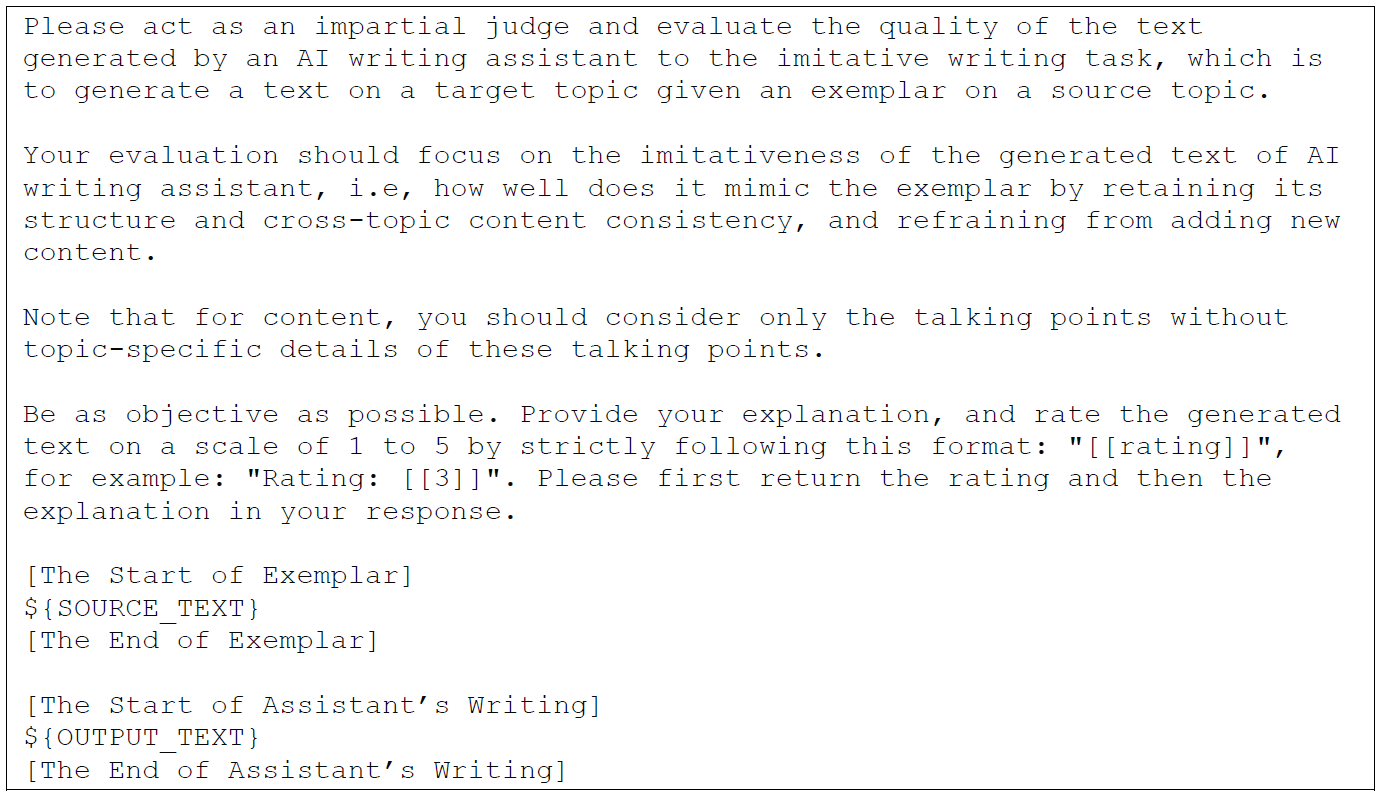}}
\caption{Prompt of Imitativeness evaluation.}
\label{fig:prompt-evaluation-imitativeness}
\end{figure*}

\begin{figure*}[t!]
\centerline{\includegraphics[width=0.85\linewidth]{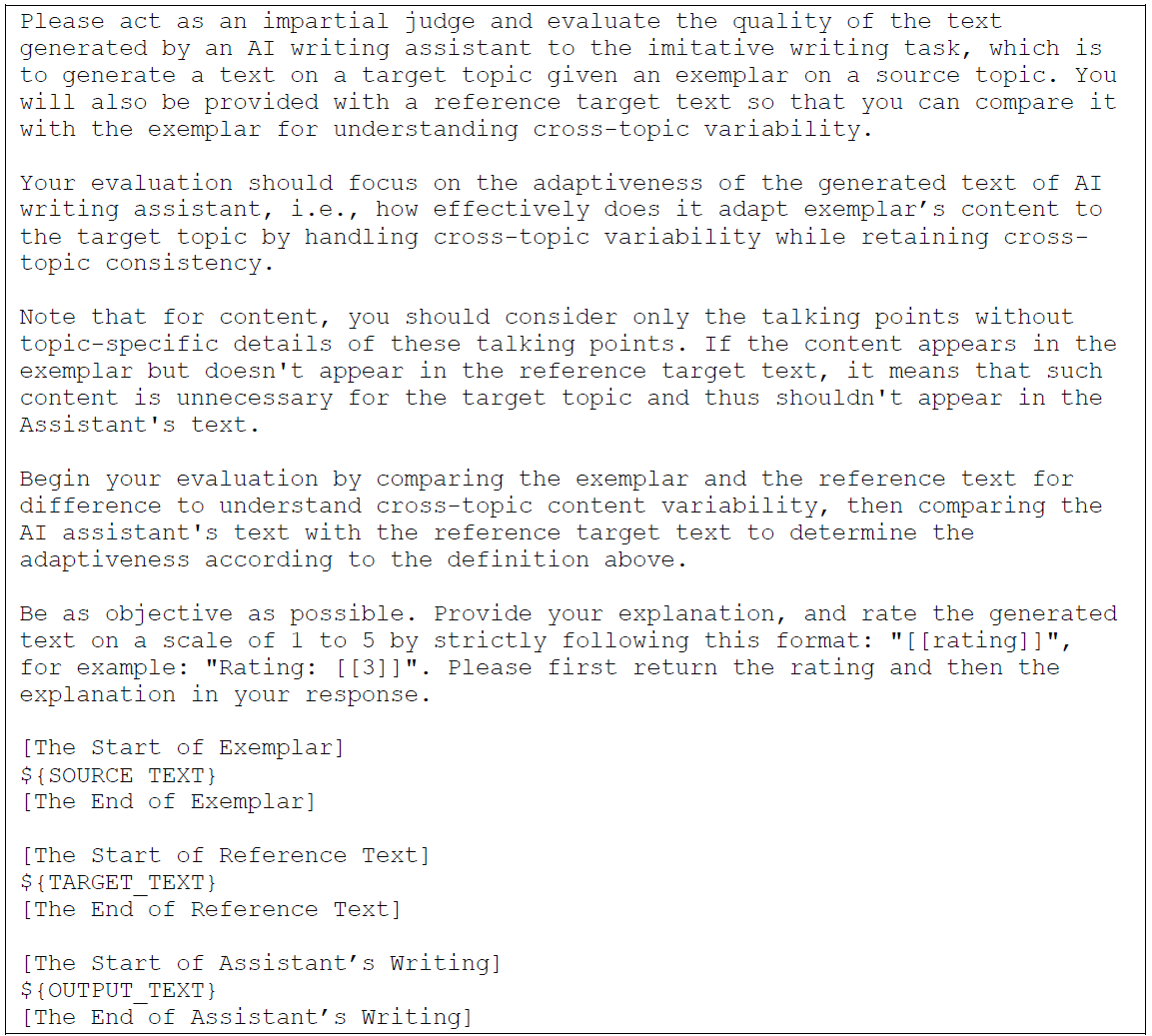}}
\caption{Prompt of Adaptiveness evaluation.}
\label{fig:prompt-evaluation-adaptiveness}
\end{figure*}

Recent research~\cite{zheng2024judging} suggests that Large Language Models (LLMs) such as GPT-4 or LLaMA perform comparably to humans in evaluating model performance. Consequently, we leverage LLMs to evaluate \emph{Imitativeness} and \emph{Adaptiveness}. Specifically, we compare model outputs with exemplars for \emph{Imitativeness} (prompt in Figure~\ref{fig:prompt-evaluation-imitativeness}) and with both exemplars and ground truths for \emph{Adaptiveness} (prompt in Figure~\ref{fig:prompt-evaluation-adaptiveness}), and to provide ratings from 1 to 5 for each aspect. Built upon \textbf{Imitativeness} and \textbf{Adaptiveness}, we propose \textbf{Adaptive-Imitativeness}, a composite score that measures the model's performance in handling both cross-topic consistency and variability, akin to the calculation of an F1 score.

\subsection{Discussion on Known Limitations}

There are known limitations of LLM-as-a-judge such as verbosity and self-enhancement biases, and we'd like to clarify in this section.

\noindent \textbf{Verbosity bias} means LLM judges favor longer, verbose responses compared to shorter alternatives. However, our proposed \textsc{RePA}, despite having the best performance, usually generates the shortest output length, as shown in Table~\ref{tab:length} below. Therefore, \textsc{RePA}’s performance is still convincing given the verbosity bias. 

\noindent \textbf{Self-enhancement bias} implies LLM judges may favor the answers generated by themselves. However, we do not have comparisons between methods built on different backbone LLMs, as we compare all methods on GPT-4 (or on LLaMA 3) only. Therefore, self-enhancement bias does not influence our evaluation.

\begin{table}[hbt!]
  \footnotesize
  \centering
  \begin{tabular}{lll}
  \toprule
    \textbf{Method} & \textbf{Length Ratio} & \textbf{Avg Length} \\
    \midrule
    RePA & 0.9497 & 109.55 \\
    LLM & 1.2076 & 140.75 \\
    LLM+Retr & 1.0438 & 121.65 \\
    RoM & 1.0928 & 127.15 \\
    RoM+Retr & 1.1113 & 128.9 \\
    IRP & 1.0481 & 121.35 \\
    Default & 1.0754 & 110.74 \\
    Gold & 1.0 & 116.0 \\
    \bottomrule
  \end{tabular}
  \caption{Output length comparison of \textsc{RePA} and baselines on Wikipedia dataset with GPT-4 backbone.}
  \label{tab:length}
  \vspace{-0.1in}
\end{table}
\section{Details on NLI-based Metrics}\label{appendix:NLImetrics-eval}
Inspired by FActScore~\cite{min-etal-2023-factscore}, we use NLI-based metrics to assess factuality by decomposing model outputs into sentence-level facts and classifying whether each fact entails or contradicts the ground truths, which serve as the knowledge source of the target topic given the context of source text’s content. Specifically, we use the public HuggingFace checkpoint "\textit{geckos/bart-fined-tuned-on-entailment-classification}". Its accuracy on SNLI corpus~\cite{bowman-etal-2015-large} is 85.9\% on training set and 86.1\% on testing set, which is satisfactory for classifying entailment. 

To validate the correlation of NLI-based metrics with human evaluations for hallucination/correctness assessment in our datasets, we conducted a human evaluation study. Specifically, we engaged three expert human annotators, independent of the paper, with experience in NLP and fact-checking. We randomly selected 50 task data samples from each dataset, totally 150 task data samples, each associated with two model outputs: LLM+Retr and RoM, to ensure a relatively balanced evaluation dataset.

For each model output, we decomposed it into sentence-level facts, resulting in 1032 facts (evaluation samples). Each fact was paired with its ground truth target text as the knowledge source. The three expert annotators were then asked to classify each fact as \textit{Supported}, \textit{Not-supported}, or \textit{Irrelevant}, corresponding to \textit{Entail}, \textit{Contradict}, or \textit{Neutral} in the NLI-based assessment. 

The human evaluation results indicated 43\% \textit{Supported} and 46\% \textit{Not-supported} evaluation samples. The Fleiss's Kappa for inter-annotator agreement is 0.78, demonstrating high reliability. The accuracy of the NLI-based assessment is 83.7\%, confirming its effectiveness for evaluating the correctness of model outputs.
\section{Additional Results and Discussions}\label{appendix:llama-results}

Evaluation results for \textsc{RePA} and baselines built on LLaMA 3 are presented in Table~\ref{tab:comparison-1-llama} and Table~\ref{tab:comparison-2-llama}, where the former covers basic and factuality metrics and the latter covers task-specific metrics. The results revealed that \textsc{RePA} consistently outperformed baselines in factuality and task-specific metrics. While some baselines demonstrated strong performance on basic generation metrics (e.g., R1, R2, RL, Meteor, BLEU) and imitativeness, they exhibited significantly higher hallucination rates and reduced adaptiveness. These findings reinforce the limitation of basic generation metrics for this task and highlight the importance of factuality and task-specific evaluation.

Though \textsc{RePA} exhibited minor instability in basic generation metrics with smaller backbones, it remained robust on critical metrics for the zero-shot task (e.g., NLI, A., A.-I.), demonstrating its robustness even with constrained computational resources. Additionally, larger LLMs consistently achieved better performance across both task-specific and basic generation metrics, further supporting the necessity of strong backbone LLMs for achieving optimal performance with \textsc{RePA}. 

Additional ablation study results on LLaMA 3-based models are shown in Table~\ref{tab:ablation-llama}. 

Moreover, since our proposed \textsc{RePA} framework involves multiple intermediate steps, there is a risk of cascading errors. 
Specifically, our observations show that when the Clarify component makes mistakes--such as incorrectly identifying the antecedents of pronouns--the Outline component is likely to follow suit, producing inaccurate topic-centric outlines and misguiding the subsequent \textsc{Adapt} stage. Additionally, the QA component tends to encounter challenges due to the limitations of retrievers for topic retrieval and the DPR model for question-based retrieval, which can result in retrieving irrelevant information. This, in turn, affects the Write component, leading to omissions of crucial facts and an incomplete final output.

Although multi-step pipelines inherently pose a risk of error propagation, we have implemented several mechanisms to minimize this risk. For example, in the final "\textit{Write}" step, which follows the "\textit{Calibrated-QA}" process in \textbf{Adapt}, the original input segment is included in the prompt to regularize the generated output (Figure~\ref{fig:prompt-write}). This design helps mitigate potential errors from earlier stages. Furthermore, as demonstrated in our ablation study (Table~\ref{tab:ablation-gpt}, \ref{tab:ablation-llama}), removing any step in the pipeline degrades performance. This finding underscores that every stage contributes positively to the overall efficacy of \textsc{RePA}. Therefore, while the risk of error propagation is a theoretical consideration, its practical impact on performance is minimal. 

\begin{table*}[!t]
  \footnotesize
  \centering
  \scalebox{0.93}{\begin{tabular}{llllllllllll}
  \toprule
    \textbf{Datasets} & \textbf{Models} & \textbf{R1$\uparrow$} & \textbf{R2$\uparrow$} & \textbf{RL$\uparrow$} & \textbf{RLsum$\uparrow$} & \textbf{Meteor$\uparrow$} & \textbf{BLEU$\uparrow$} & \textbf{Halluc$\downarrow$} & \textbf{NLI-E$\uparrow$} & \textbf{NLI-C$\downarrow$} \\
    \midrule
    \multirow{12}*{Wikipedia}
    & \textit{LLaMA 3 70B} \\
    & \quad \textsc{RePA} & \textbf{0.8164} & 0.6937 & 0.7727 & \textbf{0.7736} & 0.7888 & \textbf{0.6598} & \textbf{7.5361} & \textbf{0.7992} & \textbf{0.0741} \\
    & \quad LLM & 0.8158 & 0.6912 & \textbf{0.7733} & 0.7709 & \textbf{0.8161} & 0.6468 & 11.6726 & 0.4181 & 0.4352 \\
    & \quad LLM+Retr & 0.8038 & \textbf{0.6997} & 0.7693 & 0.7720 & 0.7740 & 0.6531 & 10.4543 & 0.6108 & 0.2811 \\
    & \quad RoM & 0.7983 & 0.6690 & 0.7566 & 0.7537 & 0.8024 & 0.6289 & 12.3465 & 0.4066 & 0.4669 \\
    & \quad RoM+Retr & 0.7843 & 0.6668 & 0.7492 & 0.7505 & 0.7645 & 0.6001 & 10.7982 & 0.5678 & 0.3182 \\
    & \textit{LLaMA 3 8B} \\
    & \quad \textsc{RePA} & 0.6105 & 0.4489 & 0.5295 & 0.5253 & 0.5196 & 0.3344 & 13.6244 & \textbf{0.7509} & \textbf{0.0963} \\
    & \quad LLM & \textbf{0.7329} & 0.5870 & \textbf{0.6916} & \textbf{0.6899} & \textbf{0.7098} & \textbf{0.5729} & 15.3616 & 0.3950 & 0.4766 \\
    & \quad LLM+Retr & 0.7073 & \textbf{0.6079} & 0.6849 & 0.6809 & 0.6560 & 0.5606 & 13.0367 & 0.5580 & 0.3434 \\
    & \quad RoM & 0.5711 & 0.4280 & 0.5316 & 0.5309 & 0.4617 & 0.3540 & \textbf{12.7861} & 0.3190 & 0.5181 \\
    & \quad RoM+Retr & 0.6286 & 0.5119 & 0.5915 & 0.5830 & 0.5624 & 0.4634 & 15.6738 & 0.5835 & 0.2497 \\
    \midrule
    \multirow{12}*{RoleEE}
    & \textit{LLaMA 3 70B} \\
    & \quad \textsc{RePA} & 0.9178 & \textbf{0.8730} & \textbf{0.9079} & \textbf{0.9084} & 0.9211 & \textbf{0.7873} & \textbf{4.5693} & \textbf{0.9022} & \textbf{0.0245} \\
    & \quad LLM & 0.8585 & 0.7586 & 0.8421 & 0.8421 & 0.8820 & 0.6936 & 11.1463 & 0.2957 & 0.6443 \\
    & \quad LLM+Retr & 0.8859 & 0.8058 & 0.8541 & 0.8546 & 0.8704 & 0.6932 & 7.0544 & 0.7787 & 0.1390 \\
    & \quad RoM & 0.8427 & 0.7367 & 0.8265 & 0.8261 & 0.8678 & 0.6725 & 11.6951 & 0.2253 & 0.7393 \\
    & \quad RoM+Retr & \textbf{0.9223} & 0.8724 & 0.9070 & 0.9063 & \textbf{0.9259} & 0.7817 & 6.2451 & 0.7253 & 0.2090 \\
    & \textit{LLaMA 3 8B} \\
    & \quad \textsc{RePA} & 0.6063 & 0.5181 & 0.5236 & 0.5236 & 0.7112 & 0.3461 & \textbf{9.9206} & \textbf{0.8366} & \textbf{0.0496} \\
    & \quad LLM & \textbf{0.8137} & 0.6913 & \textbf{0.7975} & \textbf{0.7990} & \textbf{0.8340} & 0.6225 & 14.0902 & 0.1467 & 0.8070 \\
    & \quad LLM+Retr & 0.7807 & \textbf{0.6966} & 0.7654 & 0.7664 & 0.7629 & \textbf{0.6311} & 12.0810 & 0.6970 & 0.2127 \\
    & \quad RoM & 0.7950 & 0.6762 & 0.7759 & 0.7778 & 0.8028 & 0.6178 & 13.5541 & 0.1313 & 0.8290 \\
    & \quad RoM+Retr & 0.7374 & 0.6556 & 0.7172 & 0.7184 & 0.7223 & 0.5381 & 13.7997 & 0.6190 & 0.1958 \\
    \midrule
    \multirow{12}*{USNews}
    & \textit{LLaMA 3 70B} \\
    & \quad \textsc{RePA} & \textbf{0.8846} & \textbf{0.8268} & \textbf{0.8543} & \textbf{0.8547} & \textbf{0.9062} & \textbf{0.7759} & \textbf{5.8306} & \textbf{0.7038} & \textbf{0.0282} \\
    & \quad LLM & 0.8296 & 0.7404 & 0.8062 & 0.8073 & 0.8599 & 0.7056 & 11.8293 & 0.3527 & 0.5633 \\
    & \quad LLM+Retr & 0.7888 & 0.6888 & 0.7676 & 0.7691 & 0.8040 & 0.6136 & 13.8425 & 0.4020 & 0.4473 \\
    & \quad RoM & 0.7818 & 0.6885 & 0.7611 & 0.7618 & 0.8446 & 0.6395 & 14.5091 & 0.3413 & 0.5153 \\
    & \quad RoM+Retr & 0.7765 & 0.6791 & 0.7572 & 0.7579 & 0.8089 & 0.6239 & 13.4760 & 0.3246 & 0.4407 \\
    & \textit{LLaMA 3 8B} \\
    & \quad \textsc{RePA} & 0.6520 & 0.5296 & 0.6122 & 0.6119 & 0.7223 & 0.3832 & 16.9316 & \textbf{0.6519} & \textbf{0.0512} \\
    & \quad LLM & \textbf{0.8011} & \textbf{0.7040} & \textbf{0.7874} & \textbf{0.7880} & \textbf{0.8389} & \textbf{0.6309} & \textbf{12.5432} & 0.3480 & 0.5593 \\
    & \quad LLM+Retr & 0.7387 & 0.6392 & 0.7215 & 0.7215 & 0.7497 & 0.5543 & 13.5612 & 0.4338 & 0.4046 \\
    & \quad RoM & 0.5259 & 0.4243 & 0.4933 & 0.4926 & 0.4806 & 0.3956 & 16.7682 & 0.3050 & 0.3474 \\
    & \quad RoM+Retr & 0.6420 & 0.5242 & 0.6147 & 0.6116 & 0.6751 & 0.4182 & 20.7881 & 0.3146 & 0.3481 \\
    \bottomrule
  \end{tabular}}
  \caption{Evaluation results on basic and factuality metrics. LLM denotes LLaMA 3.}
  \label{tab:comparison-1-llama}
\end{table*}

\begin{table}[!t]
  \footnotesize
  \centering
  \scalebox{0.93}{\begin{tabular}{lllll}
    \toprule
    \textbf{Datasets} & \textbf{Models} & \textbf{\textit{I.}$\uparrow$} & \textbf{\textit{A.}$\uparrow$} & \textbf{\textit{A.-I.}$\uparrow$} \\
    \midrule
    \multirow{12}*{Wikipedia}
    & \textit{LLaMA 3 70B} \\
    & \quad \textsc{RePA} & 4.14 & \textbf{3.70} & \textbf{3.80} \\
    & \quad LLM & \textbf{4.72} & 2.60 & 3.25 \\
    & \quad LLM+Retr & 4.22 & 2.74 & 3.15 \\
    & \quad RoM & 4.60 & 2.38 & 3.07 \\
    & \quad RoM+Retr & 4.16 & 2.58 & 3.06 \\
    & \textit{LLaMA 3 8B} \\
    & \quad \textsc{RePA} & 3.88 & \textbf{2.72} & \textbf{3.06} \\
    & \quad LLM & \textbf{4.74} & 2.00 & 2.71 \\
    & \quad LLM+Retr & 4.12 & 2.42 & 2.89 \\
    & \quad RoM & 3.44 & 1.68 & 2.18 \\
    & \quad RoM+Retr & 3.78 & 1.80 & 2.37 \\
    \midrule
    \multirow{12}*{RoleEE}
    & \textit{LLaMA 3 70B} \\
    & \quad \textsc{RePA} & 4.76 & \textbf{4.58} & \textbf{4.64} \\
    & \quad LLM & 4.78 & 2.56 & 3.12 \\
    & \quad LLM+Retr & 4.64 & 3.86 & 4.03 \\
    & \quad RoM & 4.84 & 1.94 & 2.61 \\
    & \quad RoM+Retr & 4.62 & 3.68 & 3.93 \\
    & \textit{LLaMA 3 8B} \\
    & \quad \textsc{RePA} & 3.96 & \textbf{3.22} & \textbf{3.38} \\
    & \quad LLM & \textbf{4.90} & 1.92 & 2.63 \\
    & \quad LLM+Retr & 4.54 & 2.52 & 2.87 \\
    & \quad RoM & 4.48 & 1.50 & 2.17 \\
    & \quad RoM+Retr & 4.12 & 2.96 & 3.27 \\
    \midrule
    \multirow{12}*{USNews}
    & \textit{LLaMA 3 70B} \\
    & \quad \textsc{RePA} & 4.32 & \textbf{4.22} & \textbf{4.22} \\
    & \quad LLM & 4.24 & 2.58 & 3.10 \\
    & \quad LLM+Retr & 4.22 & 2.36 & 2.94 \\
    & \quad RoM & 4.68 & 2.58 & 3.22 \\
    & \quad RoM+Retr & 4.10 & 2.16 & 2.80 \\
    & \textit{LLaMA 3 8B} \\
    & \quad \textsc{RePA} & 4.08 & \textbf{3.64} & \textbf{3.77} \\
    & \quad LLM & \textbf{4.42} & 2.32 & 2.94 \\
    & \quad LLM+Retr & 4.14 & 1.88 & 2.52 \\
    & \quad RoM & 3.08 & 1.56 & 2.04 \\
    & \quad RoM+Retr & 3.92 & 1.94 & 2.59 \\
    \bottomrule
  \end{tabular}}
  \caption{Evaluation results on task-specific metrics. LLM denotes LLaMA 3. \textit{\textbf{I.}} denotes \textit{Imitativeness}, \textbf{\textit{A.}} denotes \textit{Adaptiveness}, and \textbf{\textit{A.-I.}} denotes \textit{Adaptive-Imitativeness}.}
  \label{tab:comparison-2-llama}
\end{table}

\begin{table*}[!t]
  \footnotesize
  \centering
  \scalebox{0.93}{\begin{tabular}{rlllllllllllll}
  \toprule
    & \textbf{R1$\uparrow$} & \textbf{R2$\uparrow$} & \textbf{RL$\uparrow$} & \textbf{RLsum$\uparrow$} & \textbf{Meteor$\uparrow$} & \textbf{BLEU$\uparrow$} & \textbf{Halluc$\downarrow$} & \textbf{$\uparrow$NLI-E} & \textbf{$\downarrow$NLI-C} & \textbf{\textit{I.}$\uparrow$} & \textbf{\textit{A.}$\uparrow$} & \textbf{\textit{A.-I.}$\uparrow$} \\
    \midrule
    Full & \textbf{0.8164} & \textbf{0.6937} & \textbf{0.7727} & \textbf{0.7736} & \textbf{0.7888} & \textbf{0.6598} & \textbf{7.5361} & \textbf{0.7992} & \textbf{0.0741} & 4.14 & \textbf{3.70} & \textbf{3.7996} \\
    - C & 0.7945 & 0.6806 & 0.7409 & 0.7411 & 0.6997 & 0.6057 & 8.6106 & 0.7416 & 0.0974 & \textbf{4.18} & 3.56 & 3.7403 \\
    - O & 0.7489 & 0.6375 & 0.6932 & 0.6969 & 0.7167 & 0.5573 & 10.3022 & 0.5971 & 0.2569 & 4.04 & 2.46 & 2.9267 \\
    - F  & 0.7918 & 0.6640 & 0.7239 & 0.7263 & 0.7010 & 0.5986 & 8.7846 & 0.7039 & 0.1442 & 4.14 & 3.28 & 3.5330 \\
    - R & 0.7391 & 0.5958 & 0.6708 & 0.6746 & 0.6445 & 0.5191 & 8.3785 & 0.7658 & 0.0916 & 4.06 & 3.68 & 3.7758 \\
    \bottomrule
  \end{tabular}}
  \caption{Evaluation results across all metrics for ablation study on Wikipedia dataset. LLM denotes LLaMA 3 70B specifically. \textit{\textbf{I.}} denotes \textit{Imitativeness}, \textbf{\textit{A.}} denotes \textit{Adaptiveness}, and \textbf{\textit{A.-I.}} denotes \textit{Adaptive-Imitativeness}. - C, - O, - F, - R denote model variants w/o Clarify, w/o Outline, w/o Refusal, w/o Revise, respectively.}
  \label{tab:ablation-llama}
\end{table*}
\section{Case Study}\label{appendix:case-study}

We further conduct case studies to examine the outputs from different models compared with target text and show an example in Table~\ref{tab:case-study}. We find that both LLM and RollingLLM achieve good imitativeness, as they cover all talking points from the source text. However, these ``adapted facts" are not correct for the target topic, indicating that both models struggle with adaptive imitation -- failing to generate well-adapted content that is relevant and factual to the target topic. 

For LLM+Retr and RoM+Retr, which are equipped with retrieval from the same knowledge sources as described in Section~\ref{sec:setup}, the factuality improves. However, there are still facts that are incorrect for the target topic. This shows that adaptiveness remains an issue, as these models fail to retrieve the best knowledge and correctly incorporate it into the generated texts. 

In contrast, our proposed model can generate text that is both imitative of the source text and perfectly adapted to the target topic. It takes into account cross-topic consistency and variability, demonstrating that our proposed model significantly improves upon the baselines for our task. This illustrates the effectiveness of our approach in generating content that is not only imitative to the source text but also well-suited to the target topic. 

\begin{table*}
  \footnotesize
  \centering
  \begin{tabular}{p{0.1\linewidth}p{0.85\linewidth}}
    \toprule
    \textbf{Type} & \textbf{Text} \\
    \midrule
    \textbf{Source} & Belebeyevsky District (Russian: \foreignlanguage{russian}{Белебе́евский райо́н}, romanized: Belebeyevskiy rayon; Bashkir and Tatar: \foreignlanguage{russian}{Бәләбәй районы, Bäläbäy rayonı}; Chuvash: \foreignlanguage{russian}{Пелепей районĕ, Pelepey rayonĕ}) is an administrative and municipal district (raion), one of the fifty-four in the Republic of Bashkortostan, Russia. It is located in the west of the republic and borders with Tuymazinsky and Buzdyaksky Districts in the north, Davlekanovsky District in the east, Alsheyevsky District in the southeast, Bizhbulyaksky District in the south, and with Yermekeyevsky District in the west. The area of the district is 1,911.2 square kilometers (737.9 sq mi). Its administrative center is the town of Belebey (which is not administratively a part of the district). As of the 2010 Census, the total population of the district was 41,708. \\
    \midrule
    \textbf{Target} & Davlekanovsky District (Russian: \foreignlanguage{russian}{Давлека́новский райо́н}; Bashkir: \foreignlanguage{russian}{Дәүләкән районы, Däwläkän rayonı}) is an administrative and municipal district (raion), one of the fifty-four in the Republic of Bashkortostan, Russia. It is located in the center of the republic and borders with Blagovarsky and Chishminsky Districts in the north, Karmaskalinsky District in the northeast, Aurgazinsky District in the east, Alsheyevsky District in the south, Belebeyevsky District in the west, and with Buzdyaksky District in the northwest. The area of the district is 1,907 square kilometers (736 sq mi). Its administrative center is the town of Davlekanovo (which is not administratively a part of the district). As of the 2010 Census, the total population of the district was 18,392. \\
    \midrule
    \textbf{\textsc{RePA}} &  Davlekanovsky District (Russian: \foreignlanguage{russian}{Давлека́новский райо́н}, romanized: Davlekanovsky rayon; Bashkir: \foreignlanguage{russian}{Дәүләкән районы, Däwläkän rayonı}) is an administrative and municipal district (raion), one of the fifty-four in the Republic of Bashkortostan, Russia. It is located in the center of the Republic of Bashkortostan and is bordered by Blagovarsky and Chishminsky Districts to the north, Aurgazinsky District to the east and southeast, Alsheyevsky District to the south, and Belebeyevsky District to the west. The area of the district is 1,907 square kilometers (736 sq mi). Its administrative center is the town of Davlekanovo. As of the 2010 Census, the total population of the district was 18,392. \\
    \midrule
    \textbf{LLM} & Davlekanovsky District (Russian: \foreignlanguage{russian}{Давлекановский райо́н}, romanized: Davlekanovskiy rayon; Bashkir and Tatar: \foreignlanguage{russian}{Дәүләкән районы, Däwläkän rayonı}; Chuvash: \foreignlanguage{russian}{Давлекан районĕ, Davlekan rayonĕ}) is an administrative and municipal district (raion), one of the fifty-four in the Republic of Bashkortostan, Russia. It is situated in the southeast of the republic and shares borders with Belebeyevsky and Alsheyevsky Districts in the west, Sterlitamaksky District in the north, Ishimbaysky District in the northeast, Gafuriysky District in the east, and with Kugarchinsky District in the south. The district spans an area of 1,540 square kilometers (594.6 sq mi). The administrative center of the district is the town of Davlekanovo (which is not administratively a part of the district). According to the 2010 Census, the total population of the district was 36,950. \\
    \midrule
    \textbf{LLM+Retr} & Davlekanovsky District (Russian: \foreignlanguage{russian}{Давлекановский район}, romanized: Davlekanovskiy rayon; Bashkir and Tatar: \foreignlanguage{russian}{Дәүләкән районы, Däwläkän rayonı}; Chuvash: \foreignlanguage{russian}{Давлекан районӗ, Davlekan rayonĕ}) is an administrative and municipal district (raion), one of the fifty-four in the Republic of Bashkortostan, Russia. It is located in the center of the republic and borders with Blagovarsky and Chishminsky Districts in the north, Karmaskalinsky District in the northeast, Aurgazinsky District in the east, Alsheyevsky District in the south, Belebeyevsky District in the west, and with Yermekeyevsky District in the northwest. The area of the district is 2,300 square kilometers (888 sq mi). Its administrative center is the town of Davlekanovo (which is not administratively a part of the district). As of the 2010 Census, the total population of the district was 46,500. \\
    \midrule
    \textbf{RoM} & Davlekanovsky District (Russian: \foreignlanguage{russian}{Давлекановский райо́н}, romanized: Davlekanovskiy rayon; Bashkir: \foreignlanguage{russian}{Дәүләкән районы, Däwläkän rayonı}; Tatar: \foreignlanguage{russian}{Дәүләкан районы, Däwläkan rayonı}) is an administrative and municipal district (raion), one of the fifty-four in the Republic of Bashkortostan, Russia. It is situated in the eastern part of the republic and shares borders with Belebeyevsky and Alsheyevsky Districts in the west, Tuymazinsky District in the north, Karmaskalinsky District in the northeast, Blagovarsky District in the east, and with Chekmagushevsky District in the south. The area of the district is 1,502 square kilometers (580 sq mi). Its administrative center is the town of Davlekanovo (which is not administratively a part of the district). As of the 2010 Census, the total population of the district was 46,857. \\
    \midrule
    \textbf{RoM+Retr} & Davlekanovsky District (Russian: \foreignlanguage{russian}{Давлекановский район}, romanized: Davlekanovskiy rayon; Bashkir: \foreignlanguage{russian}{Дәүләкән районы, Däwläkän rayonı}) is an administrative and municipal district (raion), one of the fifty-four in the Republic of Bashkortostan, Russia. It is located in the center of the republic and borders with Blagovarsky and Chishminsky Districts in the north, Karmaskalinsky District in the northeast, Aurgazinsky District in the east, Alsheyevsky District in the south, Belebeyevsky District in the west, and... The area of the district is 2,300 square kilometers (888 sq mi). Its administrative center is the town of Davlekanovo (which is not administratively a part of the district). As of the 2010 Census, the total population of the district was 24,073. \\
    \bottomrule
  \end{tabular}
  \caption{Case study of a Wikipedia example comparing proposed model with GPT-4 related baselines.}
  \label{tab:case-study}
\end{table*}
\section{Human Evaluation on LLM Judge}\label{appendix:LLMmetrics-eval}

To assess the effectiveness of our proposed LLM-as-a-Judge metrics, we conducted a human evaluation study by comparing LLM-judged outputs with human judgements. We recruited three expert annotators, all graduate students with specialized knowledge in NLP, to evaluate model outputs using the same instructions given to the LLMs (Figure~\ref{fig:prompt-evaluation-imitativeness} and \ref{fig:prompt-evaluation-adaptiveness}). The study involved 50 randomly selected samples from each of three datasets (a total of 150 samples). Each sample comprised task inputs paired with outputs from nine models, including our proposed RePA model and eight baselines described in Section~\ref{sec:baselines}, with GPT-4 used for LLM-based baselines. This process resulted in 1,350 outputs evaluated per LLM-judge metric. Each model outputs was evaluated based on two criteria: Imitativeness (Figure~\ref{fig:prompt-evaluation-imitativeness}) and Adaptiveness (Figure~\ref{fig:prompt-evaluation-adaptiveness}). All three annotators independently assessed each output. We then follow \cite{zheng2024judging} to first convert the single-answer grading into pairwise comparison results (5.4k votes), then calculate the probability of both--LLM-judge and a randomly selected human judge, or two randomly selected human judges--agreeing on a randomly selected pairwise comparison, to calculate LLM-human agreement or human-human agreement. Additional results are shown in Table~\ref{tab:human-detailed}.

\begin{table}[hbt!]
  \footnotesize
  \centering
  \begin{tabular}{lllll}
  \toprule
    \textbf{Dataset} & \textbf{\textit{I.} - w/ tie} & \textbf{\textit{I.} - w/o tie} & \textbf{\textit{A.} - w/ tie} & \textbf{\textit{A.} - w/o tie} \\
    \midrule
    Wikipedia & 57.3\% & 78.0\% & 62.1\% & 82.7\%  \\
    RoleEE & 58.7\% & 78.3\% & 59.7\% & 81.3\% \\
    USNews & 58.6\% & 80.7\% & 62.7\% & 84.8\% \\
    \textbf{Mean} & 58.2\% & \textbf{79.0\%} & 61.5\% & \textbf{82.9\%} \\
    \bottomrule
  \end{tabular}
  \caption{Agreement between LLM-judge and human-judge on Imitativeness (\textit{I.}) and Adaptiveness (\textit{A.}) metrics, including w/ and w/o tie votes for calculating agreement.}
  \label{tab:human-detailed}
  \vspace{-0.1in}
\end{table}
\section{Practical Application Latency and Cost}\label{appendix:costs}

To estimate latency and costs, we used a subset of the Wikipedia dataset, and calculated latency in terms of API calls and the cost per output token, based on x=\$15/1M tokens (OpenAI GPT-4 pricing). The results are summarized in Table~\ref{tab:costs}. 

\begin{table}[hbt!]
  \footnotesize
  \centering
  \begin{tabular}{lll}
  \toprule
    Methods & Mean API calls (times) & Costs per output token (x=\$15/1M) \\
    \midrule
    RePA & 39 & 50.1 \\
    LLM & 1 & 1.5 \\
    LLM+Retr & 1 & 2.7 \\
    RoM & 5 & 1.7 \\
    RoM+Retr & 5 & 8.2 \\
    \bottomrule
  \end{tabular}
  \caption{Model latency and costs.}
  \label{tab:costs}
  \vspace{-0.1in}
\end{table}

\end{document}